%% file: main.tex
\documentclass[10pt]{article}
\usepackage[margin=0.92in]{geometry}
\usepackage[T1]{fontenc}
\usepackage{mathptmx}
\usepackage[scaled=0.9]{helvet}
\usepackage{microtype}
\usepackage{graphicx}
\usepackage{booktabs}
\usepackage{amsmath}
\usepackage{xcolor}
\usepackage[numbers,sort&compress]{natbib}
\usepackage[colorlinks=true,linkcolor=blue!55!black,citecolor=blue!55!black,urlcolor=blue!55!black]{hyperref}
\usepackage{caption}
\usepackage{placeins}
\usepackage{array}
\usepackage{float}
\usepackage{multirow}
\usepackage{enumitem}
\usepackage{needspace}
\setlist{nosep,leftmargin=1.4em}
\usepackage{titlesec}
\titlespacing*{\section}{0pt}{9pt plus 2pt}{4pt}
\titlespacing*{\subsection}{0pt}{7pt plus 2pt}{3pt}
\titlespacing*{\paragraph}{0pt}{5pt plus 1pt}{0.6em}
\makeatletter
\newcommand{\V}[1]{\ifcsname val@#1\endcsname\csname val@#1\endcsname\else\textcolor{red}{[\detokenize{#1}]}\fi}
\makeatother
\input{generated_an/numbers.tex}
\newcommand{\keepends}{KeepEnds}
\newcommand{\ours}{Interface-Full}
\newcommand{\legacyinput}[1]{{\let\ours\keepends\input{#1}}}
\newcommand{\kp}{k'}

\title{\textbf{Where Prefill Expert Reduction Hurts MoE Models:\\ A Few Prompt Tokens Carry Most of the Short-Answer F1 Loss}}
\author{Zhiyuan Wang\\Independent Researcher\\\texttt{zhiyuanwang@sjtu.edu.cn} \and Han Wang\\Independent Researcher\\\texttt{han.7www@gmail.com} \and Shuo Zhang\\Independent Researcher\\\texttt{zhangshuoabby@163.com}}
\date{}

\hypersetup{pdftitle={Where Prefill Expert Reduction Hurts MoE Models: A Few Prompt Tokens Carry Most of the Short-Answer F1 Loss}}
\hypersetup{pdfauthor={Zhiyuan Wang, Han Wang, Shuo Zhang}}
\begin{document}
\maketitle
\vspace{-1.6em}

\begin{abstract}
\noindent
Running prompt tokens through fewer routed experts during prefill makes mixture-of-experts (MoE) serving cheaper,
but what it damages has not been measured. We locate the damage with token-, layer- and attention-path
interventions in four MoE families on short-answer question answering. At half the experts the F1 loss is mostly
answer \emph{form}: the model still finds the answer but no longer answers briefly. The loss is made at a few
tokens: the two chat-template tokens before the answer in Qwen3-30B-A3B, and the first 8--32 tokens, where the
instruction sits, in DeepSeek-V2-Lite. Keeping them at full experts removes most of the F1 loss and about half and
a fifth of the increase in answer NLL. Reduction changes the first-token sink statistics by less than 4\%, and
restoring super experts or massive activations recovers at most a quarter of the loss. We then test placement on real
prompts with about 4\% of tokens at full experts. Putting them on the task interface keeps F1 within one point of
the unmodified model, while the same budget scattered in the documents is about 3 points worse in both prespecified
primary tests of this comparison. Time to first token falls about 14\%
on 11K-token prompts.
\end{abstract}

\section{Introduction}
\label{sec:intro}

Long retrieval-augmented and agentic prompts make prefill a large part of the time to first token (TTFT). In MoE
models \citep{yang2025qwen,deepseekai2024deepseek,muennighoff2024olmoe} much of the prefill compute is in the expert
layers: \V{flops.share.2048}\% of the multiply-adds of a 2K-token prefill of Qwen3-30B-A3B and
\V{flops.share.8192}\% at 8K tokens. Since prompt tokens mostly build the context that later tokens read, it is
tempting to give them fewer experts. ExFold \citep{wu2026exfold} does so for every prompt token and folds the
excluded experts' contribution into the retained ones with calibrated scalar projections. SlimWise
\citep{park2026slimwise} argues that pruning prefill experts trades quality for little throughput, and prunes only
during decoding. This paper measures what such reduction loses, at which tokens, and what may carry it.

Section~\ref{sec:form} splits F1 into precision, recall and answer length, and finds that at half the experts the F1
loss is mostly form in three of four families. Sections~\ref{sec:where}--\ref{sec:families} keep single tokens or
spans at full experts; in Qwen3, DeepSeek and Qwen1.5 a few tokens at one end of the prompt produce the loss.
Section~\ref{sec:mechanism} uses attention-path blocking and layer repair in Qwen3. The results are consistent with
these tokens carrying the instruction to the answer, much as safety decisions in dense chat models are anchored in
template tokens \citep{leong2025template} and the refusal direction is read at post-instruction positions
\citep{arditi2024refusal}. Keeping the prompt ends full resembles StreamingLLM \citep{xiao2023efficient}, and expert
compression can remove the super experts that create attention sinks \citep{su2025unveiling}. We therefore also
measure directly whether reduction disturbs sinks, massive activations or super experts. Section~\ref{sec:q2} tests
the consequence on real prompts, comparing placements of a fixed number of full-expert tokens.

\paragraph{Contributions.}
\begin{itemize}
\item A localisation of the F1 damage of prefill expert reduction on short-answer QA: mainly answer form in three of
four families, made at the two template tokens before the answer in Qwen3-30B-A3B, within the first 8--32 tokens of
DeepSeek-V2-Lite's standard layout and at the last two tokens in Qwen1.5-MoE; in Qwen3 it follows whatever tokens
precede the answer, not one template (Section~\ref{sec:analysis}).
\item A mechanism account for Qwen3 from path-blocking and layer-repair experiments, and prespecified direct
measurements in two families showing that the first-token sink statistics are unchanged and that super experts and
massive activations account for at most a quarter of the loss (Section~\ref{sec:mechanism}).
\item Equal-budget tests on real prompts with prespecified primary comparisons, which turn the localisation into a
placement principle: a small full-expert budget on the task interface avoids the F1 loss, and the same budget
scattered in the documents does not (Section~\ref{sec:q2}).
\end{itemize}

\section{Setting}
\label{sec:background}

\paragraph{Prefill expert reduction.}
In each MoE layer a router scores $E$ experts per token and the token runs its $K$ highest-scoring ones
($E{=}128$, $K{=}8$ in Qwen3-30B-A3B). We write $K\to\kp$ for a reduced token that keeps its $\kp$ highest-weight
experts with unchanged weights. Generated tokens always run $K$, and shared experts are never reduced. The last
prompt position computes the logits of the first answer token, so reducing it also reduces the first decoding step.
\emph{Uniform} reduction applies $K\to\kp$ to every prompt token; the \emph{unmodified} model runs $K$ everywhere; a
rule may keep some prompt tokens \emph{full} (all $K$ experts). Unless stated, reduction halves the experts ($8\to4$
for Qwen3, $6\to3$ for DeepSeek); single-token experiments use $8\to2$, where single tokens have measurable effects,
and say so.

\paragraph{Models, tasks and pipelines.}
Qwen3-30B-A3B (reasoning disabled) is the main model; DeepSeek-V2-Lite-Chat (64 routed experts, top-6, two shared
experts), Qwen1.5-MoE-A2.7B-Chat and OLMoE-1B-7B-Instruct are the other families. The analysis task is HotpotQA
\citep{yang2018hotpotqa} in the distractor setting: an instruction to give only the short answer, two
demonstrations, ten paragraphs and the question (about 1,800 prompt tokens), scored by F1 with greedy decoding.
Token, layer and attention-edge interventions run in a HuggingFace pipeline that reproduces the engine's results;
equal-budget comparisons run in vLLM \citep{kwon2023efficient} (Appendix~\ref{app:setup}). Differences are paired per
question with their reference and carry 95\% bootstrap intervals over questions. Loss sizes differ between batches
and sample sizes; each number names its batch where it matters, and Appendix Table~\ref{tab:evidence} lists the
source of every claim.

\paragraph{Prespecification.}
For four batches---three quality-only placement batches (Section~\ref{sec:q2}) and the direct-measurement batch
(Section~\ref{sec:sinks})---the prompt manifest, the conditions, the primary and secondary comparisons and their
decision rules were written into the batch's analysis script before the formal run; we call these comparisons
\emph{prespecified}. The joint quality-and-latency batch and the analysis experiments of Sections~\ref{sec:analysis}
and~\ref{sec:carry} are exploratory. There is no external registry, later batches were designed knowing earlier
results, and comparisons added after seeing the data are marked \emph{post hoc}. Primary comparisons used a two-point
margin, except necessity comparisons, which used a threshold of half of the interface rule's gain over uniform reduction: 2 points
(half the joint batch's gain) in the negative-control batch and 1.30 (half the same batch's gain) in the placement
batch. Secondary comparisons were tests against zero with Holm correction. Section~\ref{sec:q2} also uses a
one-point non-inferiority margin. It is the resolution of the joint batch's full-budget control and is stricter than the prespecified
two points; like the margin of the earlier vLLM 0.10.2 batches, it was adopted after seeing the data.

\section{What Is Lost, and Where}
\label{sec:analysis}

The interventions in this section use unmarked templates and their own baselines; Section~\ref{sec:q2} turns to
real marked prompts. Section~\ref{sec:form} uses \V{form.n} HotpotQA questions, on which the unmodified Qwen3 scores
F1 \V{form.base.f1}; the span experiments of Section~\ref{sec:where} and the family batch of
Table~\ref{tab:families} are separate 2{,}000-question runs, in which the same uniform $8\to4$ costs \V{tail.k4.n0}
and \V{fam.q3.loc.all}, and single-token experiments use 500 questions.

\subsection{At half the experts, most of the F1 loss is form}
\label{sec:form}

\begin{figure}[!tbp]
\centering
\includegraphics[width=\textwidth]{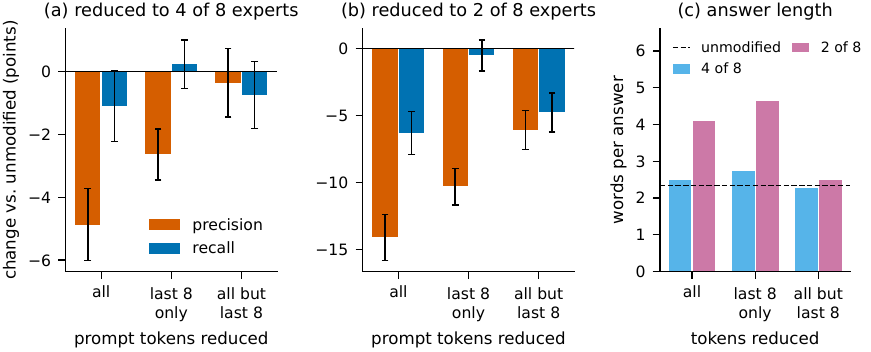}
\caption{\textbf{Form and content separate.} Qwen3-30B-A3B, HotpotQA, \V{form.n} questions; change against the
unmodified model. (a,\,b) Reducing every prompt token mostly lowers precision; reducing only the last eight tokens
lowers \emph{only} precision; reducing all but the last eight lowers recall and precision alike. (c) The answer lengthens exactly
when the last tokens are reduced.}
\label{fig:form}
\end{figure}

F1 conflates two failures, so we split it into the precision and recall of the gold answer's words and track the
answer length (Figure~\ref{fig:form}). With uniform $8\to4$, recall changes by \V{form.k4.direct_k4.recall} points
and precision by \V{form.k4.direct_k4.precision}: the answer is found nearly as often as before, but the model
writes more around it (\V{form.k4.direct_k4.words} words against \V{form.base.words} on average). The larger part of
what is lost is compliance with the instruction to answer briefly; about a point of containment is content
(Appendix~\ref{sec:content}). At $8\to2$ both fall, and the losses separate by place. Reducing \emph{only} the last
eight prompt tokens costs \V{form.k2.diag_inv8_k2.precision.abs} points of precision and
\V{form.k2.diag_inv8_k2.recall.abs} of recall, and answers become twice as long. Reducing everything \emph{but}
those eight tokens leaves the length intact and costs both recall (\V{form.k2.abl_t8_k2.recall}) and precision
(\V{form.k2.abl_t8_k2.precision}). GSM8K \citep{cobbe2021training} shows the form loss too. In the earlier unmarked
batch, uniform reduction to two experts loses \V{budget.gsm.k2.none.abs} points, and only
\V{budget.gsm.k2.none.marker}\% of outputs still end with the required ``\#\#\#\#'' marker; with the last two tokens
kept full, \V{budget.gsm.k2.t2.marker}\% do. Because F1 is length-sensitive and the instruction asks for a short
answer, ``form'' here means compliance with that request. By answer NLL the damage is less concentrated
(Section~\ref{sec:sinks}).

\subsection{In Qwen3 the form is decided at the two tokens before the answer}
\label{sec:where}

\begin{figure}[!tbp]
\centering
\includegraphics[width=\textwidth]{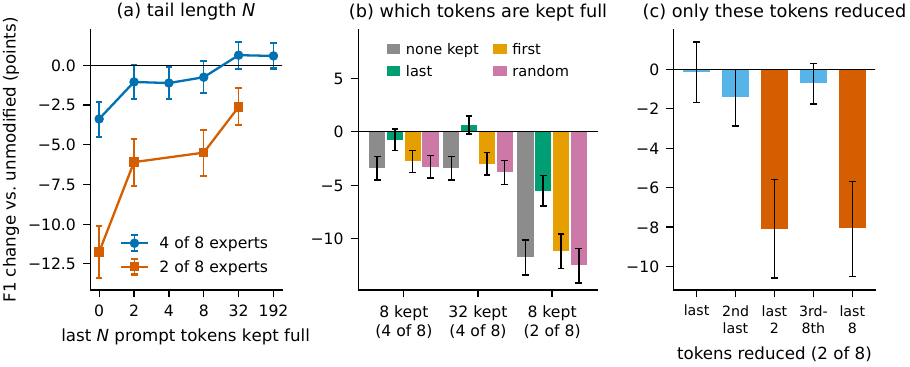}
\caption{\textbf{Which tokens need their experts} (Qwen3-30B-A3B, HotpotQA F1 against the unmodified model). (a)
Keeping the last $N$ prompt tokens full: the two-token boundary accounts for much of the format effect; a 32-token tail preserves the question as well. (b) The same number of full tokens placed
last, first, or at random on this template. (c) Reducing only the named tokens to two experts: the damage of the
last eight is the damage of the last two, which are redundant with each other.}
\label{fig:where}
\end{figure}

The damage is not spread over the prompt (Figure~\ref{fig:where}). At $8\to4$, keeping the last two tokens full and
reducing the rest moves the F1 loss from \V{tail.k4.n0} to \V{tail.k4.n2} \V{tail.k4.n2.ci}, and a tail of 32
changes F1 by \V{tail.k4.n32.full}. In the family batch the last two give \V{fam.q3.loc.keeptail2.full}, an interval
touching zero. Eight full tokens at the start (\V{budget.r_k4_t0_h8}) or at random positions
(\V{budget.r_k4_t0_x8}) recover little. We also reduced tokens in isolation at $8\to2$ (\V{only.last2.n}
questions). The last eight cost \V{only.last8.f1.abs} F1 points, the last two alone cost the same
(\V{only.last2.f1}), and the six before them cost \V{only.last8_minus_last2.f1.abs}. The two back each other up:
reduced singly, the last costs \V{only.last1.f1.abs} and the second-to-last \V{only.last2_minus_last1.f1.abs}.
Reducing the last position alone, which computes the first decoding step, therefore costs little. Because of the
backup, this does not show how much of the joint effect runs through the last position's own computation of the
first answer token (Section~\ref{sec:sinks}). With reasoning disabled a Qwen3
prompt ends in
\texttt{<|im\_start|>assistant\textbackslash n<think>\textbackslash n\textbackslash n</think>\textbackslash n\textbackslash n}:
the two tokens are the closing tag of the empty reasoning block and the blank line after it, which carry no task
content as inputs.

\textbf{In Qwen3 the sensitive tokens are those immediately before the answer, in every setting we tested} (Appendix~\ref{sec:modes}). With a base model and no
template, where the prompt ends in ``Answer:'', reducing the last two tokens at $8\to2$ costs \V{modes.raw.f1.abs} F1
points, again mostly precision. With reasoning enabled the tokens before the answer are generated, run all experts,
and reducing the prompt end has no detectable effect (\V{modes.think.f1.full}); replaying the same reasoning trace as
a prompt makes its last two tokens reducible, and reducing them costs \V{modes.trace.f1.abs} points.

\subsection{Other families: one end or the other}
\label{sec:families}

Table~\ref{tab:families} summarises the same interventions at half of each model's experts (details in Appendix
Table~\ref{tab:spans} and Figure~\ref{fig:families}). In DeepSeek-V2-Lite the form loss is made at the other end.
Reducing only the first 20 tokens reproduces it (\V{fam.ds.loc.first20}, family batch), while the sequence-start
token alone is harmless (\V{fam.ds.loc.first1.full}). Apart from the chat prefix ``User:'' and two newlines, most of those 20 tokens are the instruction, ``Answer
the question using the documents. Give only the short answer.'' DeepSeek's tail acts in the opposite direction to
Qwen3's: reducing only the last two tokens \emph{shortens} answers and raises F1 by \V{fam.ds.loc.tail2.abs}. A
vLLM prefix sweep on 400 questions (uniform loss $6.79$ in that batch) kept the first 1, 8 or 32 tokens full. Against
uniform reduction these recover $-0.42$, $+3.80$ [$+1.88$, $+5.88$] and $+8.33$ [$+5.63$, $+11.21$] points; a later
run on the same manifest adds $-0.07$ for 2 tokens and $+1.39$ [$-0.36$, $+3.13$] for 4 (Appendix~\ref{app:sink}).
We then moved the instruction, placing ten to 300 words of unrelated text before it or putting it between the
documents. Uniform reduction then shows no detectable loss ($-1.9$ to $+1.8$; vLLM, 400 questions of the placement
batch; the ten-word interval [$-4.88$, $+2.01$] is wide). In DeepSeek's standard layout we therefore cannot tell
whether the sensitivity follows the instruction or the prompt start. Qwen1.5-MoE loses content rather than form: its
answers get shorter and F1 rises, but the gold answer is found less often. Reducing only its last two tokens
reproduces its containment loss, and keeping them full leaves about a quarter of it (Appendix Table~\ref{tab:spans}).
OLMoE's loss is spread over both ends. With 32 tokens kept full at each end, none of the four models shows a
detectable HotpotQA loss in that batch; Qwen1.5's F1 rises by 1.9 and its containment is unchanged (this is not an
equivalence test).

\begin{table}[!tbp]
\centering\small
\caption{\textbf{Where the loss is made, by family} (HotpotQA, half the experts, earlier unmarked batch; change
against the unmodified model with 95\% intervals; DeepSeek's 8--32 comes from the vLLM 0.31 prefix sweep of
Section~\ref{sec:families}). Qwen1.5's loss is measured by containment (whether the gold answer
appears), since its F1 rises.}
\label{tab:families}
\setlength{\tabcolsep}{4pt}
\resizebox{\textwidth}{!}{\begin{tabular}{@{}lllll@{}}
\toprule
Model & Uniform reduction & What is lost & Where it is made & First and last 32 full \\
\midrule
Qwen3-30B-A3B & F1 \V{fam.q3.loc.all.full} & form & last two tokens & \V{fam.q3.loc.keepends.full} \\
DeepSeek-V2-Lite & F1 \V{fam.ds.loc.all.full} & form & first 8--32 tokens (instruction or prompt start) & \V{fam.ds.loc.keepends.full} \\
Qwen1.5-MoE-A2.7B & containment \V{fam.q15.loc.all.hit} (F1 \V{fam.q15.loc.all}) & content & last two tokens & containment \V{fam.q15.loc.keepends.hit.full} \\
OLMoE-1B-7B & F1 \V{fam.olmoe.loc.all.full} & form (precision) & both ends, beyond the last two & \V{fam.olmoe.loc.keepends.full} \\
\bottomrule
\end{tabular}}
\end{table}

\section{Why These Tokens}
\label{sec:mechanism}

\subsection{In Qwen3 the boundary is consistent with carrying the instruction to the answer}
\label{sec:carry}

\begin{figure}[!tbp]
\centering
\includegraphics[width=\textwidth]{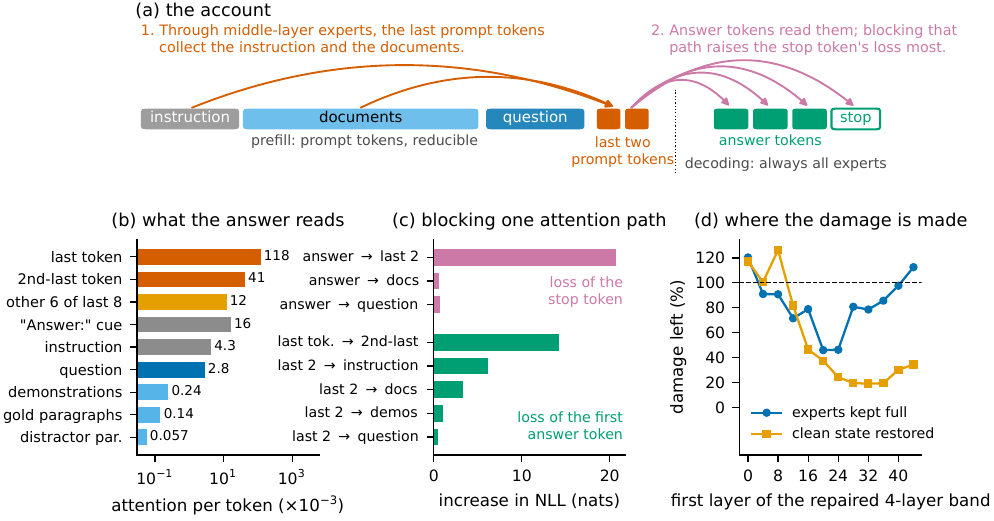}
\caption{\textbf{Why the last tokens matter in Qwen3.} (a) The account. (b) Attention each token receives from the
query that predicts the first answer token (log scale). (c) Increase in the loss of the model's own answer when one
attention path is blocked: the first answer token needs the last two prompt tokens to see the instruction and
documents; the stop token needs the answer to see those two tokens. (d) Damage left after repairing one band of
layers when the last eight tokens run two experts: the mean increase in the answer's NLL as a share of the
increase with no repair, over 300 prompts. A value above 100\% means that the partial repair made the loss
larger; the 95\% prompt-bootstrap interval of this ratio excludes 100\% for experts kept in layers 0--3 (120\%
[103, 141]) and 44--47 (113\% [105, 121]) and for the clean state restored in layers 8--11 (126\% [107, 148]).
Repairing one band mixes repaired and reduced computation, a state the model sees under neither condition. The
ratio is of means, which a few prompts dominate.}
\label{fig:mechanism}
\end{figure}

Three measurements on the unmodified model fit one account (Figure~\ref{fig:mechanism}).
\textbf{Attention.} The query that predicts the first answer token attends to the second-to-last prompt token
\V{attn.ratio.q} times more per token than to a question token. Attention does not measure importance, however: the
last token is read three times more than the second-to-last and matters far less when reduced
(Section~\ref{sec:where}). \textbf{Path blocking.} We blocked single attention paths without reducing any expert.
The first answer token depends on the last two tokens seeing the instruction (+\V{knock.ko_tail2_instr.first} nats)
and the documents (+\V{knock.ko_tail2_docs.first}), and barely on their seeing the question
(+\V{knock.ko_tail2_q.first}). The stop token depends on the answer seeing the last two prompt tokens
(+\V{knock.ko_ans_tail2.stop} nats), not the documents (+\V{knock.ko_ans_docs.stop}) or the question
(+\V{knock.ko_ans_q.stop}). Hiding as many random prompt tokens from the answer costs nothing
(Appendix~\ref{app:sink}). A block removes a heavily attended token outright. This is a far larger perturbation than
reduction, which raises the answer NLL by 1.42 nats in the 200-question batch of Section~\ref{sec:sinks}, so
blocking shows what flows through the boundary tokens rather than how reduction damages it. \textbf{Layer repair.}
With the last eight tokens at two experts, we restored all experts in one band of four layers at a time. Layers
20--23 and 24--27 repair the most, about half (each leaves \V{restore.band.best}\% of the damage); each band in the first twelve
or last eight layers repairs at most a tenth. Three partial repairs make the damage larger (above 100\% in
Figure~\ref{fig:mechanism}d), so mixed reduced and repaired states do not add up monotonically. These results fit
middle-layer experts writing ``how to answer'' into the boundary tokens and the answer reading it there, much as
safety decisions in dense chat models are anchored in template tokens \citep{leong2025template}. They do not
identify the mechanism uniquely.

\subsection{Reduction does not disturb the sink statistics we measured}
\label{sec:sinks}

Sinks coincide with massive activations on the first token and on delimiters
\citep{sun2024massive,gu2024attention,barbero2025why,queipo2025sinks}, and in Qwen3-30B-A3B these are produced by a
few routed \emph{super experts}, whose removal collapses the sinks \citep{su2025unveiling}. $K\to\kp$ keeps a token's
highest-weight experts, so super experts are unlikely to be removed. Qwen3's two sensitive tokens, however, are
heavily attended delimiters and could be secondary sinks. We therefore measured the question directly in a
prespecified HuggingFace batch on 200 standard-layout HotpotQA questions each for DeepSeek-V2-Lite and Qwen3-30B-A3B
(Appendix~\ref{app:sink}). In this batch uniform reduction costs $10.3$ F1 points in DeepSeek and an undetectable
$-1.06$ [$-4.41$, $+2.33$] in Qwen3. Super experts are identified as in \citet{su2025unveiling}. Each intervention is
scored by the share of the uniform-reduction increase in answer NLL (of the unmodified model's own answer) that it
removes. This score is far less noisy than F1 at this size, and Qwen3's rows rest on it alone.

\begin{table}[!tbp]
\centering\small
\caption{\textbf{Direct tests of sink-related explanations} (HuggingFace, 200 HotpotQA questions per model; DeepSeek
$6\to3$, Qwen3 $8\to4$). \emph{Change}: ratio of the statistic under uniform reduction to the unmodified model, minus
one, averaged over layers 1 onward and heads (first token: attention from the last eight prompt positions, hidden
norm, largest absolute channel; last eight prompt tokens: attention from the answer, mean norm and largest channel;
for DeepSeek these are not its sensitive tokens). \emph{Recovery}: share of the uniform-reduction increase in answer
NLL removed, $\overline{\mathrm{NLL}_{\text{unif}}-\mathrm{NLL}_{\text{cond}}}/\overline{\mathrm{NLL}_{\text{unif}}-\mathrm{NLL}_{\text{orig}}}$,
in percent; rules keep 32 (DeepSeek) or 8 (Qwen3) tokens full. \emph{Minus position}: paired difference to keeping the
sensitive positions full (post hoc, except for the two signal rules, for which it is the quantity of the prespecified
rule (h)). Restorations and random full tokens mix clean and reduced states (Section~\ref{sec:sinks}). 95\% bootstrap intervals over questions. Answer NLL rises from 1.10 to 2.64 nats
(DeepSeek) and 0.22 to 1.64 (Qwen3).}
\label{tab:sinks}
\setlength{\tabcolsep}{3.5pt}
\resizebox{\textwidth}{!}{\begin{tabular}{@{}lllll@{}}
\toprule
 & \multicolumn{2}{c}{DeepSeek-V2-Lite} & \multicolumn{2}{c}{Qwen3-30B-A3B} \\
\midrule
\multicolumn{5}{@{}l}{\emph{Is the sink disturbed? (change under uniform reduction)}} \\
First token: attention / norm / largest channel & \multicolumn{2}{l}{$+0.3$\% / $-0.2$\% / $-0.1$\%} & \multicolumn{2}{l}{$+1.5$\% / $-3.2$\% / $-3.2$\%} \\
Last eight tokens: attention / norm / largest channel & \multicolumn{2}{l}{$-4.3$\% / $-4.3$\% / $-1.4$\%} & \multicolumn{2}{l}{$-0.3$\% / $-8.3$\% / $-10.0$\%} \\
Hide first token from last eight and answer, $\Delta$NLL (random token) & \multicolumn{2}{l}{$+1.07$ ($+0.00$) nats} & \multicolumn{2}{l}{$+40.6$ ($+0.01$) nats} \\
Super experts (layer/expert); boundary selections removed & \multicolumn{2}{l}{L3/E54, L4/E38; 12.5\%} & \multicolumn{2}{l}{L1/E68, L2/E92, L3/E82; 10.8\%} \\
\midrule
 & Recovery & Minus position & Recovery & Minus position \\
\midrule
Keep the sensitive positions full (first 32 / last 8) & $+22$ [$+10$, $+33$] & --- & $+49$ [$+30$, $+65$] & --- \\
Restore super experts on every reduced token & $+1$ [$-4$, $+6$] & $-20$ [$-31$, $-9$] & $+8$ [$-11$, $+24$] & $-41$ [$-64$, $-20$] \\
Restore massive activations to clean values & $-4$ [$-9$, $+1$] & $-25$ [$-36$, $-14$] & $+6$ [$-10$, $+22$] & $-43$ [$-64$, $-22$] \\
Keep tokens full by super-expert signal & $+2$ [$-4$, $+7$] & $-19$ [$-30$, $-9$] & $+13$ [$-6$, $+30$] & $-35$ [$-60$, $-12$] \\
Keep tokens full by massive-activation signal & $+5$ [$-3$, $+12$] & $-17$ [$-30$, $-3$] & $+4$ [$-14$, $+20$] & $-44$ [$-69$, $-23$] \\
Keep random tokens full & $+6$ [$0$, $+11$] & $-16$ [$-27$, $-5$] & $-14$ [$-33$, $+2$] & $-63$ [$-86$, $-40$] \\
Restore the first token's keys and values & $0$ [$-6$, $+6$] & --- & $-17$ [$-38$, $0$] & --- \\
Restore the sensitive positions' keys and values & $+19$ [$+8$, $+30$] & --- & $-353$ [$-527$, $-247$] & --- \\
\bottomrule
\end{tabular}}
\end{table}

\paragraph{The first-token sink statistics are unchanged.} Both models have a functioning first-token sink. Hiding
it from the last eight prompt tokens and the answer raises the answer NLL by 1.1 nats in DeepSeek and 40.6 in Qwen3,
against nothing for a random token. Yet uniform reduction changes its attention mass, norm and largest activation by
at most 3.2\%. The prespecified rule (a), defined on DeepSeek, reads \emph{excluded}, and Qwen3's first-token
statistics are within the same band. We also measured Qwen3's last eight prompt tokens, which include the two
delimiters (we did not measure the two separately). The attention the answer pays them is unchanged ($-0.3$\%), and
their norm falls by 8.3\%, inside the $\pm10$\% band, so rule (b), prespecified on attention and norm, reads
excluded. Their largest channel, which is not part of the rule, falls by 10.0\% [$-10.2$, $-9.9$], just outside the
band, and restoring massive activations removes $+6$\% [$-10$, $+22$] of the loss. These statistics are layer- and
head-averaged scalars and do not capture changes in the direction of keys and values. In Qwen3, cleaning the first
token's keys and values raises the NLL ($-17$\%), which an undisturbed sink would not predict unless the mixed state
itself does harm. In DeepSeek the ten-word layout shows no detectable loss in this batch ($-2.9$ [$-7.7$, $+1.8$] F1;
HuggingFace, 200 questions), although its first-token statistics change as little as in the standard layout. These
tests do not answer whether the boundary tokens act as sink-like registers in the unmodified model. Hiding Qwen3's
last two prompt tokens from the answer raises its NLL by 23.9 nats, against 40.6 for hiding the first token from the
last eight tokens and the answer (Appendix~\ref{app:sink}).

\paragraph{Super experts and massive activations account for less than about a quarter of the loss.} $K\to\kp$
removes only 11--13\% of the boundary selections of a super expert, so by construction this reduction rarely removes
super experts. The tests therefore rule out their removal, not a disturbance of sink mechanisms through other
experts. Restoring super experts on every reduced token, or massive activations to their clean values, removes at
most 8\% of the loss in point estimate. The prespecified rules (c) and (d) read \emph{does not explain} (upper bound
below one half) in both models. The positional rule itself does not pass that threshold in NLL, so the informative
comparison is the paired one: both restorations recover 20--43 percentage points less than keeping the sensitive
positions full (post hoc). Keeping full the tokens ranked highest by super-expert or massive-activation signals also
recovers 17--44 points less than position. The prespecified rule (h) calls a signal rule tied with position if the
95\% lower bound of this difference exceeds $-25$ percentage points, and worse if the upper bound is below $-25$.
Neither holds in either model, so the rule reads undecided. DeepSeek's signal rules are also no better than random
tokens, while Qwen3's are. The 32 tokens DeepSeek's signal rules keep share one or two with its first 32, and its F1
stays at the uniform level ($-9.1$ and $-9.6$, against $+1.6$ for the first 32).

\paragraph{The answer NLL is not concentrated where F1 is.} We split the answer NLL post hoc. In Qwen3, 64\% of the
increase falls on the first answer token, which the last prompt position predicts. Keeping the last eight tokens full
removes 67\% [51, 81] of that part and 8\% [$-35$, $+42$] of the increase on later answer tokens; keeping the whole
task interface full (Section~\ref{sec:q2}) removes 94\% and 89\%. The stop token's share (7\%) is too small to
estimate. In DeepSeek the first 32 tokens remove 75\% [63, 86] of the stop token's increase, 28\% of the first answer
token's and none of the later answer tokens', yet restore F1 fully. The few boundary tokens thus account for most of
the short-answer F1 loss but not for most of the likelihood loss. In Qwen3, much of what they do account for may be
the last position's own reduced computation of the first answer token.
Separating that computation from the later answer tokens that read the boundary needs the clean-keys-and-values
condition, and it fails. Cleaning the last eight tokens' keys and values while their own states stay reduced raises
Qwen3's NLL to 6.68, far above uniform reduction. Unit tests on a small model of the same architecture confirm the
patch itself: cleaning every prompt position's keys and values with the last position kept full reproduces the
unmodified model exactly. In DeepSeek, whose cleaned tokens do not include the last position, the same condition
reproduces keeping the tokens full. In Qwen3, restorations that mix clean and reduced states are not monotone;
random full tokens and clean first-token keys and values also raise the NLL. We report every such condition
(Table~\ref{tab:sinks}) but read none of them as evidence on the boundary channel, and a net recovery from a mixed
state is not an upper bound on a quantity's contribution.

\section{Where the Full-Expert Budget Goes}
\label{sec:q2}

The localisation predicts that, at a fixed number of full-expert tokens, \emph{where} they go decides whether the
loss is avoided. We test this on real prompts in which the application marks the documents.
\emph{\ours{}} reduces only the marked documents and keeps the \emph{task interface}---system message, instruction,
demonstration questions and answers, current question and chat-template tokens---full, on average about 4\% of
tokens (0.6--26\% by task). Two controls keep, prompt by prompt, the same number of full tokens. \emph{Equal-budget
ends} puts them at the two prompt ends. The \emph{negative control} puts them at random positions inside the
documents, at least 32 tokens from either end, so that the whole interface is reduced. A third, the \emph{inverted
control}, reduces exactly the interface. Two rules need no marks: \emph{template role} reduces the user-message
content, and \keepends{} keeps a fixed 32+32 tokens. All rules share one vLLM 0.31 hook, checked on every prompt against the server's token ids
(Appendix~\ref{app:rules}). Table~\ref{tab:place} collects the comparisons from a joint quality-and-latency batch
(760 prompts) and three prespecified quality-only batches (about 2{,}100 prompts each; Appendices
\ref{app:joint}--\ref{app:interf2}).

\begin{table}[!tbp]
\centering\small
\caption{\textbf{Where the full-expert budget goes} (vLLM 0.31, F1 points, task-averaged, 95\% bootstrap intervals
over questions unless marked 90\%). Batches: \emph{joint} (760 prompts, H100, exploratory), \emph{neg.} (negative
control), \emph{interf.} (interface), \emph{place.} (placement; H200), \emph{A100} (2{,}000 long-system HotpotQA
questions). Status: \emph{pri}/\emph{sec} prespecified primary/secondary, \emph{post} post hoc; the prespecified
equivalence margins were $\pm2$. Seven placements of the negative control: one in neg., one in interf., five in
place.}
\label{tab:place}
\setlength{\tabcolsep}{4pt}
\resizebox{\textwidth}{!}{\begin{tabular}{@{}llll@{}}
\toprule
Comparison & Batch & Difference & Status \\
\midrule
\multicolumn{4}{@{}l}{\emph{Qwen3-30B-A3B, $8\to4$}} \\
Uniform $-$ unmodified & joint / neg. / place. & $-3.51$ / $-2.15$ / $-1.83$ & --- \\
\ours{} $-$ unmodified & neg. / place. & $+0.46$ [$-0.62$, $+1.57$] / $+0.76$ [$-0.49$, $+2.06$] & sec \\
 & joint & $+0.78$, 90\% [$-0.31$, $+1.89$] & post \\
\textbf{\ours{} $-$ negative control} & neg. & $\mathbf{+3.22}$ [$+1.74$, $+4.73$] & pri (rule met) \\
 & place., five placements & $\mathbf{+3.05}$ [$+1.47$, $+4.68$]; per placement $+2.33$ to $+3.88$ & pri (rule met) \\
 & interf. & $+1.63$ [$-0.08$, $+3.29$], Holm $p=0.18$ & sec \\
Negative control $-$ uniform & neg. / place. & $-0.61$ [$-1.69$, $+0.43$] / $-0.46$ [$-1.42$, $+0.45$] & post / sec \\
Inverted control $-$ unmodified & neg. & $-1.86$ [$-3.45$, $-0.33$], Holm $p=0.07$ & sec \\
\ours{} $-$ header only (header full, rest of budget in documents) & interf. & $+1.01$ [$-0.61$, $+2.64$], undecided & pri \\
\ours{} $-$ equal-budget ends & joint & $+0.63$ [$-0.54$, $+1.83$] & post \\
 & neg. / interf. / place. & 90\% [$-0.98$, $+0.41$] / [$-0.93$, $+0.65$] / [$-1.07$, $+0.45$] & post \\
\ours{} $-$ template role & joint & $+1.23$ [$-0.59$, $+3.07$] & post \\
\keepends{} (32+32) $-$ unmodified, long system prompt & A100 & $-1.55$ [$-2.50$, $-0.60$] & post \\
\midrule
\multicolumn{4}{@{}l}{\emph{DeepSeek-V2-Lite, $6\to3$}} \\
Instruction between documents: \ours{} $-$ equal-budget ends & interf. & $-1.67$ [$-3.56$, $+0.19$], undecided & pri \\
Standard layout: \ours{} $-$ negative control & interf. / place. & $+7.92$ [$+4.74$, $+11.20$] / $+8.09$ [$+5.05$, $+11.26$] & post / sec \\
\quad same, containment instead of F1 & interf. / place. & $-3.75$ [$-7.50$, $0.00$] / $-2.38$ [$-5.88$, $+1.12$] & post \\
\bottomrule
\end{tabular}}
\end{table}

\paragraph{The budget avoids the loss on the interface but not in the documents.}
Uniform $8\to4$ costs 1.8--3.5 F1 points on Qwen3 in point estimate; the $-1.83$ of the placement batch has an
interval touching zero. With the small budget on the interface, \ours{} stays within one point of the unmodified
model in the two prespecified secondary comparisons, with 95\% lower bounds of $-0.62$ and $-0.49$. These comparisons
were prespecified as tests against zero; the one-point margin is post hoc. The containment of \ours{} shows no detectable change
($+0.77$ [$-0.90$, $+2.42$] in the joint batch). Scattered at random in the documents, the same number of full tokens
helps little. The negative control is below \ours{} by $3.22$ and $3.05$ points in the two prespecified primaries,
which meets their rule of at least half of \ours{}'s gain over uniform reduction. In a prespecified secondary it is
below by $1.63$, with an interval touching zero. Where uniform reduction was run in the same batch, the negative
control is at most 0.43--0.45 points above it (95\% upper bounds). The seven placements come from two prompt sets of
2{,}098 prompts (1{,}498 questions shared) on two GPU types. Two placements on the same prompts and GPU differ by 1.8
points, so placement variance, which these intervals omit, is as large as question sampling.

The other two prespecified primaries are undecided. One compares \ours{} with keeping only the assistant header full
and moving the rest of the budget into the documents ($+1.01$ [$-0.61$, $+2.64$]); the other is DeepSeek with the
instruction between documents (below). A fifth, on DeepSeek in the placement batch, was not run because no layout
passed its screen. Reducing only the interface costs about as much as uniform reduction ($-1.86$ against $-2.15$ in
the same batch), although not significantly after Holm correction. Gold paragraphs matter more than distractors
(Appendix~\ref{sec:content}); we did not test placing the budget there. On DeepSeek the negative control is about 8
F1 points below \ours{}, but its containment is if anything higher, so the gap is likely answer length, consistent with
Section~\ref{sec:form}. \ours{} itself is not lossless everywhere. On question-first LongBench and code completion it
is 1.6--2.4 points (F1; edit similarity for code) below the unmodified model (Supplementary Table~S21,
Supplement~S3.1), and its containment drops on tool-output prompts (Supplement~S3.2); we have
not located this residual.

\paragraph{Marks, template roles and positions are not distinguished in our templates.}
In the three quality-only batches the equal-budget ends control is within $\pm1.1$ points of \ours{} (90\%
intervals), and in the joint batch template role differs by $+1.23$ [$-0.59$, $+3.07$]. The reason is that in our
templates the sensitive tokens lie at the ends these rules keep. No margin was prespecified for these comparisons.
Two of the three 90\% intervals lie within $\pm1$ and the third just outside, and the equal-budget controls take
their per-prompt budget from the marks. Keeping the boundary is also not sufficient. With a 700-token system prompt, a
fixed 32+32-token rule loses 1.55 points on 2{,}000 questions, mostly form, although it keeps the assistant header
full. The equal-budget ends control, whose head of about 300 tokens also stops short of the instruction, loses little
(Appendix~\ref{app:joint}). In Qwen3, then, form also depends on how much of a long system prompt is reduced,
through tokens we have not located. The results support a placement principle---keep the template boundary, the
system prompt, the instruction and the question full---rather than a better selector. The prespecified primary that
directly tests marks against positions away from the ends, DeepSeek with the instruction between the documents, is
uninformative: that layout shows no loss to recover ($-1.67$, undecided).

\section{What It Buys in Serving}
\label{sec:q3}

In the joint batch (one H100, short answers, TTFT one request at a time), the median TTFT of \ours{} on LongBench
prompts \citep{bai2023longbench} (median 10.9K tokens) is $0.86$ [$0.86$, $0.87$] of the unmodified model, against
0.85 for uniform reduction. On HotpotQA prompts of about 2K tokens it is $1.02$ [$1.01$, $1.03$], slightly slower.
There uniform reduction saves only 4\%, although the multiply-add model predicts far more, for reasons we have not
found. The intervals omit start-to-start variation (4\%). The same implementation is
slower than the unmodified model on Qwen3.6-35B-A3B at 1--4K tokens (Appendix~\ref{app:serving} and Supplement~S9). The saving grows with prompt length, consistent with SlimWise
\citep{park2026slimwise}; our results say which tokens to keep full when reducing.

\section{Related Work}
\label{sec:related}

\paragraph{Fewer experts at inference.}
Static methods remove experts once \citep{lu2024experts,lasby2025reap,lee2024stun,dong2025domain}; dynamic methods
skip experts per token by gate weight or frequency \citep{lu2024experts,huang2024mixture,chen2025eac,huang2025modes},
allocate them per layer \citep{chittyvenkata2025lexi}, or reduce top-$k$ uniformly; \citet{chen2026need} find that
two thirds of the chosen experts suffice, a milder setting than our halving that we did not test. ExFold
\citep{wu2026exfold} reduces every prompt token and folds the excluded experts into the retained ones with calibrated
scalar projections. SlimWise \citep{park2026slimwise} argues that pruning prefill experts costs quality for little
throughput, prunes for decoding only, and notes that accuracy can hide pruning-induced changes in generation length,
the form effect we locate. MC-MoE \citep{huang2024mixture} protects a few important tokens chosen by attention scores
(compared in Supplement~S3.6); we ask where reduction fails rather than propose a selector.
\paragraph{Sensitive tokens, sinks and super experts.}
Quantization keeps a few pivot or outlier tokens intact \citep{liu2024intactkv,chen2024prefixquant}, selected by
activation magnitude. The tokens we find are the task boundary, and by the statistics we measured they are not
outliers. Attention sinks \citep{xiao2023efficient,gu2024attention,barbero2025why} coincide with massive activations
\citep{sun2024massive,queipo2025sinks}, which in MoE models a few super experts produce \citep{su2025unveiling}. In
the two models we measured, prefill expert reduction leaves the first-token sink statistics we measured unchanged, and these
quantities explain little of the loss; we do not rule out secondary sinks at the boundary tokens.
\paragraph{Where a prompt's task information sits.}
Final positions aggregate what the output needs \citep{geva2023dissecting,hendel2023context,todd2023function}, label
words in demonstrations act as anchors \citep{wang2023label}, and context matters less once the task is recognised
\citep{sia2024context}; refusal is mediated by a direction read at post-instruction positions
\citep{arditi2024refusal} and safety decisions are anchored in template tokens \citep{leong2025template}. KV-cache and prompt reduction
\citep{li2024snapkv,fu2024lazyllm,jiang2023longllmlingua} keep or drop tokens by similar structural priors; we keep
every token and change only its expert budget.

\section{Limitations}
\label{sec:limits}

The analysis uses short-answer QA with reasoning off; GSM8K, code completion and reasoning-on runs are checks only.
The loss we localise is measured by F1, which is length-sensitive. In answer NLL the boundary accounts for about half
(Qwen3) or a fifth (DeepSeek) of the increase, and with reasoning on uniform reduction showed no detectable loss. We tested only half the experts and $8\to2$, not
milder settings, and models up to 30B parameters apart from a Qwen3.6 serving check. The mechanism is traced only in Qwen3, on unmarked templates,
with interventions that do not identify it uniquely and whose partial repairs are not monotone; the last position's
own computation of the first answer token is not separated from what later tokens read. The direct sink tests
use 200 questions and averages over layers and heads, which could hide changes in single sink heads; Qwen3's rows rest
on NLL. The equal-budget evidence is from Qwen3 and, for the negative control, DeepSeek-V2-Lite; we varied placement,
not amount; the sensitive tokens always lay at a prompt end, so marks were never tested where positions fail.
\ours{} leaves an unlocated residual of up to 2.4 points on question-first, tool and code layouts, and a fixed 32+32
rule loses form on long system prompts although it keeps the boundary. Latency comes from one GPU (TTFT one request at
a time, throughput at concurrency four) without prefix caching.

\section{Conclusion}
\label{sec:conclusion}

Halving the experts of prompt tokens in MoE prefill mostly costs answer form on short-answer QA. In F1 the cost is
made at a few tokens at one end of the prompt: in Qwen3 the template tokens before the answer, in DeepSeek-V2-Lite
the start of its standard layout, where the instruction sits. The first-token sink statistics we measured change by
less than 4\%, and super experts and massive activations account for at most about a quarter of the loss. A small
full-expert budget on the task interface avoids the F1 loss, while the same budget scattered in the documents does
not. In practice, the template boundary, the system prompt, the instruction and the question should stay at full
experts, and the documents can be reduced.

\FloatBarrier
\bibliographystyle{plainnat}
\bibliography{references}

\clearpage
\appendix
\section{Guide to the Appendices}
\label{app:guide}

These appendices support claims in the main text: the direct sink tests (Appendix~\ref{app:sink},
Section~\ref{sec:sinks}); the full account of the placement batches and serving measurements
(Appendices~\ref{app:placement} and~\ref{app:serving}, Sections~\ref{sec:q2}--\ref{sec:q3}); the rules, controls and
implementation (Appendix~\ref{app:rules}); the evidence sources and per-batch results
(Appendices~\ref{app:evidence}--\ref{app:interf2}); the further analysis of the answer boundary and of content
(Appendix~\ref{app:more-analysis}, Sections~\ref{sec:form}--\ref{sec:families}); and the setup, statistics and
reproducibility details (Appendix~\ref{app:setup}). An earlier evaluation of these rules as a method---the label-free
selector, the vLLM 0.10.2 batch, ExFold's protocol, MC-MoE, Qwen3.6 serving and the A100 rerun---is in a separate
supplement, provided with this paper as the file \texttt{supplement.pdf} and cited as Supplement~S$n$; the main text
relies on it only where it cites it.

\section{Direct Tests of Sink-Related Explanations}
\label{app:sink}

\paragraph{Batch.} One H100 PCIe (80\,GB), HuggingFace transformers with eager attention, bfloat16, greedy decoding,
reasoning off. DeepSeek-V2-Lite-Chat ($6\to3$) on 200 HotpotQA questions in the standard layout and the same 200
questions with ten words of unrelated text before the instruction (\emph{pre10}); Qwen3-30B-A3B ($8\to4$) on 200
standard-layout questions (the mechanism experiments of Section~\ref{sec:carry} use 300). A vLLM 0.31 part
adds the DeepSeek prefix-sweep points $k=2,4$ on the 400-question standard-layout manifest of the placement batch
(original outputs identical to that batch on 89.8\% of questions, below the 99\% merge threshold, so the two runs are
reported side by side, not merged). The conditions, statistics and decision rules (a)--(h) were written into the
analysis script before the formal run, after a six-question pilot.

\paragraph{Statistics.} For each question the unmodified model first generates its answer; every condition is then
scored by the summed NLL of that answer, stop token included, and by the F1 of its own greedy generation. Sink
statistics average layers 1 onward: for the first token the attention it receives from the last eight prompt
positions (head-averaged), its hidden-state norm and its largest absolute channel; for the last eight prompt tokens
the attention the answer positions pay them and their mean norm and largest channel. The last eight include Qwen3's
two delimiters; we did not compute the statistics for the two alone. A change $c$ is the ratio of
means to the unmodified model minus one; a sink is \emph{intact} if the 95\% interval of $c$ lies in $[-0.10,
+0.10]$ and \emph{broken} if $c\le-0.20$ with upper bound below zero. Recovery $R$ is defined in
Table~\ref{tab:sinks}; a restoration \emph{explains} the loss if $R\ge0.5$ with lower bound above $0.25$ and
\emph{does not} if the upper bound is below $0.5$. Massive activations follow \citet{sun2024massive} ($|h|>100$
and above 1000 times the layer median); super experts follow \citet{su2025unveiling} (99.5th percentile and a tenth of
the maximum of the experts' largest output activations, first 75\% of layers). Intervals are 4{,}000-sample
bootstraps over questions with all conditions and layouts of a question resampled together.

\paragraph{Conditions.} Uniform reduction; the positional rule (DeepSeek: first 32 tokens full; Qwen3: last 8);
restorations on top of uniform reduction---super experts on every reduced token (\emph{se\_all}) or only at the
boundary and first token (\emph{se\_bnd}), massive entries of the hidden state set to their clean values, and clean
keys and values at the boundary (DeepSeek: first 32 tokens; Qwen3: last 8), at the first token, or at as many random non-boundary tokens; signal rules that keep
full the tokens ranked highest by the number of layers that select a super expert, by the massive-activation ratio,
or at random, with the positional rule's count; and attention blocks paired with random non-boundary controls of
the same size.

Table~\ref{tab:sink-more} gives the answer NLL and F1 of every condition.

\begin{table}[!htbp]
\centering\small
\caption{\textbf{Further results of the direct-test batch.} Scores and answer NLL averaged over the 200 questions;
F1 of each condition's greedy generation.}
\label{tab:sink-more}
\setlength{\tabcolsep}{4pt}
\begin{tabular}{@{}lcccc@{}}
\toprule
 & \multicolumn{2}{c}{DeepSeek, standard} & \multicolumn{2}{c}{Qwen3} \\
Condition & NLL & F1 & NLL & F1 \\
\midrule
Unmodified & 1.10 & 49.2 & 0.22 & 67.1 \\
Uniform & 2.64 & 38.9 & 1.64 & 66.0 \\
Positional rule (first 32 / last 8) & 2.31 & 50.7 & 0.95 & 66.4 \\
\ours{} & --- & --- & 0.39 & 68.8 \\
Restore super experts, all reduced tokens & 2.62 & 37.7 & 1.53 & 65.6 \\
Restore super experts, boundary only & 2.64 & 38.9 & 1.65 & 65.6 \\
Restore massive activations & 2.70 & 39.2 & 1.56 & 67.6 \\
Clean boundary keys and values & 2.35 & 50.0 & 6.68 & 52.9 \\
Clean first-token keys and values & 2.64 & 40.8 & 1.88 & 66.6 \\
Clean random-token keys and values & 2.62 & 40.0 & 1.67 & 64.5 \\
Signal rule, super expert & 2.61 & 40.1 & 1.45 & 66.7 \\
Signal rule, massive activation & 2.57 & 39.6 & 1.58 & 66.1 \\
Signal rule, random & 2.55 & 40.3 & 1.84 & 66.6 \\
\bottomrule
\end{tabular}
\end{table}

\paragraph{DeepSeek's insensitive layout.} With ten words before the instruction, uniform reduction changes F1 by
$-2.93$ [$-7.73$, $+1.78$] against $-10.27$ [$-15.52$, $-5.29$] in the standard layout; first-token attention changes
by $+0.8$\% (pre10) and $+0.3$\% (standard) and its norm by $-0.15$\% in both, so rule (a) reads \emph{excluded}. No restoration or signal
rule moves the pre10 NLL by more than 4\% of its (small) increase.

\paragraph{Keys and values.} Because a token influences later tokens only through its keys and values, cleaning the
first 32 tokens' keys and values in DeepSeek should reproduce keeping them full, and it does (NLL 2.35 against 2.31,
F1 50.0 against 50.7); this is a consistency check, not a separate finding. In Qwen3, cleaning the last eight tokens'
keys and values while their own states stay reduced raised the answer NLL to 6.68 (recovery $-353$\% [$-527$,
$-247$]) and lowered F1 to 52.9. Here the cleaned tokens include the last position, whose own reduced state predicts
the first answer token while attending to clean keys and values of its own and its neighbours' positions, a mixed
state the model never sees. The patch itself is tested on small models of both architectures: cleaning every prompt
position's keys and values with the last position kept full reproduces the unmodified model exactly, and with the
last position reduced it reproduces every answer token after the first. We report the condition with the other
mixed-state conditions and do not interpret any of them as evidence on the boundary channel. Rule (e) reads undecided in both models (below).

\paragraph{Attention blocks.} Every boundary block is specific against its random control: in DeepSeek, hiding the
last two prompt tokens from the answer raises its NLL by $+3.13$ nats against $0.00$, hiding the first 32 tokens from
the last eight and the answer by $+16.1$ against $+0.42$; in Qwen3 the same blocks give $+23.9$ and $+31.8$ against at
most $+0.15$. In Qwen3, hiding the first 32 tokens raises the NLL less than hiding the first token alone ($+31.8$
against $+40.6$); we have not explained this.

\paragraph{Prefix sweep (vLLM).} Keeping the first 2 or 4 DeepSeek tokens full changes F1 against uniform
reduction by $-0.07$ [$-1.61$, $+1.56$] and $+1.39$ [$-0.36$, $+3.13$]; the unmodified model scores 47.15 and uniform
reduction 41.00 in this run.

\paragraph{Prespecified rules and their verdicts.}
(a) \emph{First-token sink} (DeepSeek, standard against pre10): \emph{excluded} if attention and norm are both intact
under uniform reduction in the standard layout; \emph{decoupled} or \emph{not excluded} by whether the standard layout
loses while its statistics change as much as, or more than, pre10's. Verdict: excluded.
(b) \emph{Secondary sink at the last eight tokens} (Qwen3; DeepSeek descriptive), on attention and norm only:
\emph{excluded} if both are intact; otherwise decoupled or not excluded by whether keeping the last eight full
restores them together with F1. Verdict: excluded (the largest channel, $-10.0$\% [$-10.2$, $-9.9$], is outside the
band but was not part of the rule).
(c) \emph{Massive activations} and (d) \emph{super experts}: \emph{explains} if restoring them recovers at least half
(lower bound above a quarter), \emph{does not explain} if the upper bound is below half. Verdicts: does not explain in
both models, for super experts also with restoration at the boundary only.
(e) \emph{Boundary keys and values} (supporting): the boundary carries the loss through its keys and values if clean
boundary keys and values recover more than clean random ones (lower bound of the difference above zero) and at least
half; not specific if the interval of the difference contains zero; otherwise undecided. Verdict: undecided in both
(DeepSeek: difference $+0.17$ [$+0.06$, $+0.29$] but recovery below half; Qwen3: difference $-3.52$ [$-5.21$, $-2.46$]).
(f) \emph{Attention blocks}: specific if the boundary block raises NLL more than its random control (lower bound
above zero). Verdict: every block specific in both.
(g) \emph{Prefix sweep} (vLLM, DeepSeek): descriptive only, with the earlier sweep points merged if at least 99\% of
unmodified outputs match; they did not (89.8\%), so the points are reported side by side.
(h) \emph{Signal rules}: effective if recovery exceeds random tokens' (lower bound of the difference above zero); tied
with position if the lower bound of the difference to the positional rule exceeds $-25$ percentage points; worse if its
upper bound is below $-25$; otherwise undecided. Verdict: undecided in both (DeepSeek: not better than random tokens; Qwen3: better
than random, with margins to position of $-35$ [$-60$, $-12$] and $-44$ [$-69$, $-23$] points).
The other paired differences to the positional rule in Table~\ref{tab:sinks} are post hoc.

\paragraph{Where the NLL increase lies (post hoc).} Split into the first answer token, later answer tokens and the stop
token, the increase under uniform reduction is 42\% / 45\% / 13\% in DeepSeek and 64\% / 29\% / 7\% in Qwen3. Keeping
DeepSeek's first 32 tokens full removes $+28$\% [$+8$, $+46$], $0$\% [$-15$, $+13$] and $+75$\% [$+63$, $+86$] of the
three parts; keeping Qwen3's last eight full removes $+67$\% [$+51$, $+81$], $+8$\% [$-35$, $+42$] and $+47$\% [$-132$,
$+120$], and keeping its whole task interface full $+94$\%, $+89$\% and $+25$\%. Restoring super experts or massive
activations removes at most $+17$\% of any part, with intervals including zero.

\paragraph{Medians (post hoc).} Recovery is a ratio of means that a few questions dominate. Over the questions whose
answer NLL rises by more than 0.1 nats (149 in DeepSeek, 78 in Qwen3), the median per-question recovery is 42\% and
83\% for the positional rules, 1\% and 11\% for restoring super experts, $-3$\% and 18\% for restoring massive
activations, 3\% and 34\% and 4\% and 18\% for the super-expert and massive-activation signal rules, and 2\% and 3\%
for random tokens.

\FloatBarrier

\needspace{12\baselineskip}\section{Placement Batches: Full Account}
\label{app:placement}

This appendix gives the full account of the comparisons summarised in Section~\ref{sec:q2}.

\emph{Claim tested: at the number of full tokens per prompt that \ours{} sets (we do not vary the amount), where the
full tokens go decides whether the loss of uniform reduction is avoided on short-answer QA; and, more finely, which
part of the task interface has to stay full.}
The interventions of Section~\ref{sec:analysis} used unmarked templates; they explain the loss of uniform reduction
but not where the residual loss of \ours{} on marked prompts comes from, which we have not localised.

\paragraph{Joint batch.}
One H100 SXM (80\,GB) serves Qwen3-30B-A3B with vLLM 0.31 (bfloat16, greedy decoding, reasoning off, 8192-token
step budget, at most four concurrent sequences). One frozen manifest of 760 real marked prompts is used by every
method: 200 HotpotQA \citep{yang2018hotpotqa} questions with a $\sim$700-token system prompt ahead of the instruction, 200 in the standard
layout, and 60 from each of six LongBench \citep{bai2023longbench} question-answering sets, truncated in the middle to at most 15{,}000
tokens (mean 5{,}968 prompt tokens over all 760 prompts; median 1.8K and 2.3K for the two HotpotQA layouts and 10.9K for LongBench). Every method starts its own server. Cycle 1 runs all prompts with full
generation; cycles 2 and 3 restart every server, reverse the order of methods in cycle 2, and run a fixed
128-prompt timing subset. TTFT and end-to-end latency are measured one request at a time and throughput at concurrency four. Quality
is F1 per question, averaged per task and then over the eight tasks; intervals are 95\% bootstrap intervals over
questions, resampled within tasks, with the two HotpotQA layouts (the same 200 questions) resampled together;
$\geq$ marks non-inferiority at a one-point margin (90\% lower bound above $-1$), and no method meets equivalence
(90\% interval within $\pm1$). These calls are neither prespecified nor corrected across methods, and the margin is about the batch's resolution
(the full-budget control's 90\% half-width).

\begin{table}[!tbp]
\centering\small
\caption{\textbf{Quality and latency in one batch} (Qwen3-30B-A3B, vLLM 0.31, one H100, 760 marked prompts,
$8\to4$ on reduced tokens). \emph{Slots}: routed-expert work relative to the unmodified model. \emph{Quality}: F1
change against the unmodified model on the marked prompts (task-averaged, 95\% interval; a positive value means a higher score than the reference row, so the unmarked prompt scores 0.39 above the marked one). \emph{90\%}: the 90\% interval that decides non-inferiority ($\geq$: lower bound above $-1$). \emph{\ours{} minus}:
the same-question difference between \ours{} and that method. Latency columns: median per-prompt ratio to the unmodified model in the same cycle. \emph{760}: cycle 1, all
prompts, the same prompts as the quality column. \emph{128}: cycles 2 / 3 (restarted servers) on a fixed timing
subset of 16 prompts per task, which over-weights long LongBench prompts (median prompt 7.6K tokens in the subset against 2.6K over all 760
prompts); TTFT and end-to-end latency are measured one request at a time, throughput at concurrency four. Full cycle-1 ratios are in Appendix~\ref{app:joint}.}
\label{tab:joint}
\setlength{\tabcolsep}{3.4pt}
\resizebox{\textwidth}{!}{\begin{tabular}{@{}lcccccccc@{}}
\toprule
Method & Slots & Quality & 90\% & \ours{} minus & TTFT, 760 & TTFT, 128 & End-to-end, 128 & Throughput, 128 \\
\midrule
Unmodified (reference) & 1.00 & 0 & & & 1 & 1 / 1 & 1 / 1 & 1 / 1 \\
Unmodified, unmarked prompt & 1.00 & $+0.39^{\geq}$ {\scriptsize [$-0.89$, $+1.78$]} & {\scriptsize [$-0.69$, $+1.53$]} & & 0.99 & 0.99 / 1.01 & 1.00 / 1.01 & 1.00 / 1.00  \\
Full-budget control & 1.00 & $+0.38^{\geq}$ {\scriptsize [$-0.82$, $+1.65$]} & {\scriptsize [$-0.61$, $+1.44$]} & & 1.03 & 1.04 / 1.05 & 1.53 / 1.57 & 0.91 / 0.92  \\
\midrule
Uniform & 0.50 & $\mathbf{-3.51}$ {\scriptsize [$-5.67$, $-1.26$]} & {\scriptsize [$-5.29$, $-1.63$]} & $\mathbf{+4.28}$ {\scriptsize [$+2.11$, $+6.50$]} & 0.89 & 0.86 / 0.86 & 0.87 / 0.87 & 1.22 / 1.22  \\
\keepends{} (32, 32) & 0.51 & $-0.37$ {\scriptsize [$-1.89$, $+1.12$]} & {\scriptsize [$-1.68$, $+0.91$]} & $+1.15$ {\scriptsize [$-0.09$, $+2.42$]} & 0.93 & 0.87 / 0.87 & 0.88 / 0.87 & 1.21 / 1.21  \\
Template role & 0.51 & $-0.45$ {\scriptsize [$-2.64$, $+1.82$]} & {\scriptsize [$-2.29$, $+1.42$]} & $+1.23$ {\scriptsize [$-0.59$, $+3.07$]} & 0.95 & 0.87 / 0.88 & 0.88 / 0.88 & 1.22 / 1.20  \\
Equal budget, ends & 0.52 & $+0.15$ {\scriptsize [$-1.47$, $+1.78$]} & {\scriptsize [$-1.25$, $+1.53$]} & $+0.63$ {\scriptsize [$-0.54$, $+1.83$]} & 0.96 & 0.88 / 0.88 & 0.89 / 0.89 & 1.22 / 1.21  \\
Equal budget, rand & 0.52 & $-0.13$ {\scriptsize [$-1.79$, $+1.49$]} & {\scriptsize [$-1.55$, $+1.25$]} & $+0.91$ {\scriptsize [$-0.25$, $+2.13$]} & 0.94 & 0.88 / 0.88 & 0.88 / 0.89 & 1.22 / 1.20  \\
Equal budget, span & 0.52 & $+0.46^{\geq}$ {\scriptsize [$-1.03$, $+1.99$]} & {\scriptsize [$-0.76$, $+1.73$]} & $+0.31$ {\scriptsize [$-0.75$, $+1.37$]} & 0.90 & 0.89 / 0.88 & 0.89 / 0.89 & 1.21 / 1.20  \\
\ours{} & 0.52 & $+0.78^{\geq}$ {\scriptsize [$-0.53$, $+2.15$]} & {\scriptsize [$-0.31$, $+1.89$]} & & 0.93 & 0.87 / 0.88 & 0.88 / 0.89 & 1.21 / 1.21  \\
\bottomrule
\end{tabular}}
\end{table}

\paragraph{Keeping the interface or the prompt ends full recovers the loss of uniform reduction.}
Table~\ref{tab:joint} shows that uniform $8\to4$ loses 3.5 F1 points and \ours{} is $+4.3$ points above it on the
same questions; every rule that keeps the prompt ends or the marked interface full shows no detectable loss, so this
batch does not tell interface from ends. \ours{} is non-inferior to the unmodified model at the one-point margin (90\% interval
$[-0.31, +1.89]$), but its interval is too wide to establish equivalence; so is the equal-budget span control. Its
small lead is not a form artifact of F1 that hides lost content: containment (whether the gold answer appears in the
output) changes by $+0.77$ [$-0.90$, $+2.42$] under \ours{} and by $-2.15$ [$-4.67$, $+0.33$] under uniform
reduction. The other rules show no detectable loss but do not reach non-inferiority; the marks themselves ($+0.39$) and the
full-budget code path ($+0.38$) are non-inferior without establishing equivalence. The Holm-adjusted $p$ values in
Table~\ref{tab:negctl} and Appendix Table~\ref{tab:interf} test differences from zero, not non-inferiority.

\paragraph{Moving the protected budget into the documents recovers little of the gain.}
We ran a second, quality-only batch with a negative control that keeps, prompt by prompt, exactly as many full
tokens as \ours{} but places all of them inside the marked documents and away from the prompt ends
(Section~\ref{sec:q2} and Appendix~\ref{app:rules}), and an inverted control that reduces only the tokens \ours{} keeps full. The
methods, the primary comparison and its decision rule were written into the analysis script before the batch ran
(Appendix~\ref{app:negctl}). The batch has 2{,}098 prompts on the same model, engine and GPU type: standard,
long-system and question-first HotpotQA (300 questions, the same in all three layouts; two long-system prompts were
removed for every method because the documents could not hold the negative control's budget), six LongBench QA sets
and two code-completion sets, LCC \citep{guo2023longcoder} and RepoBench-P \citep{liu2023repobench} (150 each). The primary comparison is \ours{} minus the negative control over seven QA
datasets (HotpotQA layouts pooled per question); the rule required a 95\% lower bound above zero and a point
estimate of at least two points, set at about half of \ours{}'s gain over uniform reduction in the joint batch.
\ours{} is $+3.22$ [$+1.74$, $+4.73$] above the negative control (Table~\ref{tab:negctl}), which meets it. The same
number of full tokens therefore helped where \ours{} put them and not where this placement of the negative control
put them: the negative control is $2.76$ points below the unmodified model and $-0.61$ [$-1.69$, $+0.43$] against
uniform reduction (post hoc). That is not evidence of no effect, but it bounds it: the negative control is at most $0.43$ points above uniform
reduction (95\% upper bound); as a share of \ours{}'s gain over uniform reduction ($2.61$ points here) it recovers
$-0.24$ [$-1.03$, $+0.15$] (post hoc), a ratio whose denominator is itself uncertain. \ours{} is again non-inferior to the unmodified model ($+0.46$ [$-0.62$, $+1.57$]; 90\%
lower bound $-0.43$), and its containment is unchanged ($+0.08$ [$-1.22$, $+1.33$]). This is one random placement;
seven placements in all give gains of 1.6 to 3.9 points (below).

\paragraph{What the controls identify.}
The negative control moves the budget off the instruction, question, assistant header and prompt ends at once, so
the primary shows that this set matters, not which member. Two results narrow it, neither conclusively. The inverted control, which reduces only the tokens that \ours{} keeps full
(0.6--26\% of a prompt, depending on the task), is $1.86$ points below the unmodified model (95\% interval
[$-3.45$, $-0.33$], unadjusted; Holm $p=0.07$) and $2.32$ [$0.78$, $3.81$] below \ours{} (post hoc, Table~\ref{tab:negctl}); uniform
reduction loses $2.15$ [$0.55$, $3.69$] in this batch. And on question-first HotpotQA the equal-budget ends control,
whose head of about 60 tokens covers none of the current question's tokens in any of the 300 prompts, scores like
\ours{}, but both are about one point below the unmodified model there, in both this batch and the interface batch
(Tables~\ref{tab:negctl-task} and~\ref{tab:interf-task}). This is compatible with the template tokens before the answer carrying most of the effect in these Qwen3
templates, not with the question text being free to reduce.

\begin{table}[!tbp]
\centering\small
\caption{\textbf{Negative-control batch} (Qwen3-30B-A3B, vLLM 0.31, one H100, 2{,}098 marked prompts, $8\to4$,
quality only). F1 differences on seven QA datasets, equally weighted; code is edit similarity. Intervals are
unadjusted 95\% bootstrap intervals with HotpotQA questions resampled across their three layouts; $p$ is
Holm-adjusted over the six prespecified secondary comparisons. The last three rows were added after the data were
seen and are not part of the correction.}
\label{tab:negctl}
\begin{tabular}{@{}llc@{}}
\toprule
Comparison & Difference [95\%] & Holm $p$ \\
\midrule
\textbf{\ours{} $-$ negative control (primary)} & $\mathbf{+3.22}$ [$+1.74$, $+4.73$] & --- \\
\midrule
Inverted control $-$ unmodified & $-1.86$ [$-3.45$, $-0.33$] & 0.070 \\
Negative control $-$ equal budget, ends & $-3.50$ [$-5.07$, $-1.90$] & $<$0.001 \\
Negative control $-$ unmodified & $-2.76$ [$-4.33$, $-1.21$] & 0.003 \\
\ours{} $-$ unmodified & $+0.46$ [$-0.62$, $+1.57$] & 0.63 \\
Primary, question-first HotpotQA only & $+2.01$ [$-1.18$, $+5.19$] & 0.63 \\
Primary, code completion only & $+1.60$ [$-1.00$, $+4.28$] & 0.63 \\
\midrule
Negative control $-$ uniform (post hoc) & $-0.61$ [$-1.69$, $+0.43$] & --- \\
\ours{} $-$ equal budget, ends (post hoc) & $-0.28$ [$-1.08$, $+0.53$] & --- \\
\ours{} $-$ inverted control (post hoc) & $+2.32$ [$+0.78$, $+3.81$] & --- \\
\bottomrule
\end{tabular}
\end{table}

\paragraph{More placements, a second family and the assistant header.}
Two further quality-only batches (the \emph{interface} and \emph{placement} batches, Appendices~\ref{app:interf}
and~\ref{app:interf2}) add placements of the negative control, DeepSeek-V2-Lite and direct tests of the interface.
\emph{Placements.} On Qwen3, five placements on the 2{,}098 prompts of the interface batch (a different prompt set
from the negative-control batch with the same count; 1{,}498 questions are shared) leave \ours{} $+3.05$ [$+1.47$,
$+4.68$] above the negative control averaged per question (prespecified; per placement $+2.33$ to $+3.88$), and the
negative control is $-0.46$ [$-1.42$, $+0.45$] against uniform reduction (secondary). A prespecified placement in the
interface batch gave $+1.63$ [$-0.08$, $+3.29$] (Holm $p=0.18$), not significant after Holm correction; on the shared questions the two
H100 placements differ by 1.8 points with identical unmodified and \ours{} outputs, so placement variance is as large
as question sampling and is not in these intervals. The seven estimates (two prompt sets, two GPU types) average
$2.87$ (s.d.\ $0.72$), all positive (Appendix~\ref{app:interf2}).
\emph{DeepSeek.} On the standard layout, where \ours{} and the ends control lose nothing, the negative control is
$7.92$ [$4.74$, $11.20$] (interface batch, split by layout post hoc; $4.28$ when pooled with an insensitive layout,
secondary) and $8.09$ [$5.05$, $11.26$] (placement batch, secondary) F1 points below \ours{}. This is a gap in answer
length, not content: the negative control's answers are longer, and its containment is if anything higher (\ours{}
minus the negative control $-3.75$ [$-7.50$, $0.00$] and $-2.38$ [$-5.88$, $+1.12$], post hoc). \emph{Assistant header.} On Qwen3, keeping only the
eight-token header full and the rest of the budget in the documents is $1.01$ [$-0.61$, $+2.64$] below \ours{}
(undecided); reducing the header and moving its budget into the documents is $1.86$ [$0.38$, $3.39$] below (Holm
$p=0.064$). \emph{Where DeepSeek is sensitive.} With the instruction between the documents or after 10--300 words
of unrelated text, uniform reduction shows no detectable loss ($-1.9$ to $+1.8$; the 10-word interval [$-4.88$,
$+2.01$] still allows 70\% of the standard-layout loss of $6.79$), so no layout passed the prespecified screen and
marks against positions away from the ends were not tested. On the standard layout, keeping the first token full
recovers nothing ($-0.42$ against uniform), the first 8 about half ($+3.80$ [$+1.88$, $+5.88$]) and the first 32 all
of it ($+8.33$), while keeping the last 8 full is \emph{worse} than uniform ($-4.06$ [$-6.72$, $-1.46$]), consistent
with reduced tails shortening DeepSeek's answers (Section~\ref{sec:analysis}). In the DeepSeek layouts we tested,
the interface and a prompt end coincide wherever reduction costs anything; on Qwen3 an unlocated residual remains,
with \ours{} and the ends control both one to two points below the unmodified model on question-first
and instruction-in-the-middle HotpotQA (Table~\ref{tab:interf-task}).

The direct tests of sink-related explanations are in Section~\ref{sec:sinks} and Appendix~\ref{app:sink}.

\paragraph{Marks and positions are not distinguished in these templates.}
\ours{} is ahead of \keepends{} by $+1.15$ points with an interval that touches zero, and ahead of the three
equal-budget positional controls by $+0.31$ to $+0.91$, all intervals including zero; differences of up to about two
points are not excluded in the joint batch. In the later batches the difference to the equal-budget ends control is
smaller: $-0.28$ (90\% interval [$-0.98$, $+0.41$]) in the negative-control batch and $-0.16$ (90\% [$-0.93$,
$+0.65$]) in the interface batch, both within $\pm1$ and so equivalent at a one-point margin (post hoc), and $-0.33$
(90\% [$-1.07$, $+0.45$]) in the placement batch, which is not; the three estimates are nearly the same, so the
split is a threshold effect, and the prespecified margins were $\pm2$. This concerns quality, not the tokens chosen: the masks differ (\V{flops.role.std.full}\ against 64 full tokens in
the standard layout), and we did not measure their overlap.

\paragraph{Why positional rules suffice here, and when they would not.}
All three equal-budget controls cover the prompt ends, where the sensitive tokens of Section~\ref{sec:analysis} lie
in these templates, so the role labels add nothing detectable only over positional rules given the same budget. The
template-role rule, which needs no marks and reduces the instruction and question inside the user message, is also
not distinguished from \ours{} ($+1.23$ [$-0.59$, $+3.07$]). The equal-budget controls are not label-free: their
per-prompt budget is the one the marks give \ours{}. The label-free \keepends{} ($H=N=32$) cannot cover a long system prompt and the instruction after it, and on 2{,}000
long-system HotpotQA questions it loses $3.51$ points with vLLM 0.10.2 and $1.55$ [$0.60$, $2.50$] and $1.03$ in two
vLLM 0.31 runs on an A100, where \ours{} is $1.87$ and $1.51$ ahead (Supplement~S9); the joint batch's
$-0.54$ on 200 questions is consistent with these. The marks set the budget without tuning, but so does the chat
template. The amount matters too: a head of about 300 tokens loses little on long system prompts where a 32-token head
loses form (Appendix~\ref{app:joint}), and DeepSeek's head sweep rises with length. Where
the interface is at neither end (an instruction between documents, tool outputs after the question), marks could
also place better; the five DeepSeek layouts with the instruction off the start showed no detectable loss, so this is
untested.

\section{Serving Measurements}
\label{app:serving}

\emph{Claim tested: halving the routed-expert work of prompt tokens makes serving faster.}

\paragraph{Same batch, same implementation.}
Answers in the joint batch are short (median five generated tokens), so end-to-end latency follows TTFT. Over all
760 prompts (cycle 1), the median TTFT of \ours{} is 0.93 of the unmodified model, and the saving depends on prompt
length: on the 360 LongBench prompts (median 10.9K tokens after truncation) it is 0.86 [0.86, 0.87], on the 400 HotpotQA
prompts (median about 2K tokens) 1.02 [1.01, 1.03], slightly slower. Uniform reduction gives 0.85 and 0.96 (every
rule in Appendix Table~\ref{tab:ttft-length}). These are cycle-1 medians from one server start per method; the
intervals resample prompts and omit start-to-start variation. On the long-prompt-weighted
timing subset (128 prompts, median 7.6K tokens), restarted twice, every reduced rule reaches 0.86--0.89, and
throughput at concurrency four rises by $1.20$--$1.22\times$; we did not measure throughput separately by prompt length. Between restarts the unmodified model's median TTFT varies by 4\% and its throughput by under 1\%. On longer
prompts (8K--32K tokens, one request at a time, a separate H100 PCIe session) \ours{} lowers the median TTFT of
Qwen3-30B-A3B to 0.88--0.90 of the unmodified model (Supplement~S10).

\paragraph{The implementation can erase the saving.}
The full-budget control forces the row-appending call in every engine step, including decoding steps with no
reduced token, where the rules take the plain fused call; it costs 3--5\% TTFT and makes end-to-end latency
1.5--1.7 times longer. The absent saving of \ours{} at about 2K tokens is consistent with a fixed per-step cost of the hook (not profiled),
but uniform reduction, which takes the plain fused call, also saves only 4\% there against 15--17\% on synthetic
2K-token prompts (Supplement~S9); we have not explained this. Longer outputs, other batch shapes or a model
whose expert GEMMs do not shrink can turn the saving into a loss.

\paragraph{No stable speed-up on Qwen3.6.}
On Qwen3.6-35B-A3B (a hybrid linear-attention MoE with 256 routed experts and one shared expert \citep{qwen36card}) the same row-appending approach was slower than the
unmodified model in two formal runs on real marked prompts of 1.1--3.9K tokens with one generated token: every
reduced method reached only 0.863--0.932 of its throughput (Supplement~S9). Qwen3-30B-A3B shows
no saving at these lengths either, so prompt length and model are confounded in this comparison. At 32K tokens TTFT improved by 2--8\% (2--7\% for \ours{}; Supplement~S10); at 8K and 16K it was too unstable to compare, for a reason we have not identified. One untested hypothesis is
that with 256 experts and a 4096-token step budget the halved expert inputs fall into the same padded kernel tile
(Supplement~S9--S10).

\paragraph{The cost model and the measured gain.} Amdahl's law over the expert multiply-add share predicts about
$+41\%$, $+26\%$ and $+18\%$ throughput for uniform reduction at 2K, 8K and 16K tokens. Wall time does not follow
multiply-adds (Supplement~S3.7), so this is a reference value, not a bound: with vLLM 0.31 the measured
gains are close to it and slightly above at 8K and 16K (A100, synthetic prompts: $+38\%$, $+27\%$, $+19\%$, Supplementary Table~S23), and the earlier vLLM 0.10.2 measurement is well above it
(\V{v3speed.uniform.8192.gain}\%/\V{v3speed.uniform.16384.gain}\% at 8K/16K), for reasons we have not identified.

\FloatBarrier
\section{Rules, Controls and Implementation}
\label{app:rules}

\begin{figure}[!tbp]
\centering
\includegraphics[width=\textwidth]{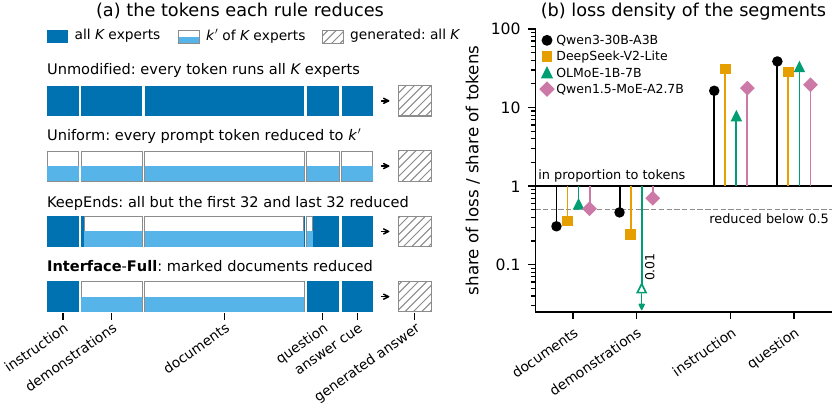}
\caption{\textbf{Prompt roles and expert budgets.} Left: the tokens each rule reduces in the illustrated prompt;
generated tokens retain all experts. Right: segment loss density measured with an exploratory label-free probe
(Supplement~S3.4; thresholds not prospectively validated), relative to segment token share; task
segments carry far more of the model's answer loss than their share of tokens. For simplicity panel (a) draws the demonstrations as wholly reduced under
\ours{}; the serving rule reduces only the documents inside them.}
\label{fig:teaser}
\end{figure}

Section~\ref{sec:background} defines prefill expert reduction and the notation; Figure~\ref{fig:teaser} illustrates the rules on one prompt.

\paragraph{Rules that keep some prompt tokens full.}
\keepends{} keeps the first and last 32 prompt tokens full. \ours{} reads spans that the application marks in the
prompt (documents, including those inside demonstrations, and GSM8K worked examples) and reduces only those; the
instruction, example questions and answers, current question and chat-template tokens stay full. No gold answers are used.
\emph{Template role} reduces the user-message content and keeps the system message and chat-template tokens full,
needing no application marks. The three \emph{equal-budget controls} keep, for each prompt, exactly as many full
tokens as \ours{} but choose them by position: all at the two ends (\emph{ends}), or the \keepends{} ends plus the
remainder at random positions (\emph{rand}) or as one contiguous span (\emph{span}). The equal-budget ends control
splits its budget evenly between head and tail. We call the tokens that \ours{} keeps full---system message,
instruction, example questions and answers, current question, and the chat-template tokens, including the
\emph{assistant header} that ends every chat prompt (in Qwen3 the eight tokens
\texttt{\textbackslash n<|im\_start|>assistant\textbackslash n<think>\textbackslash n\textbackslash n</think>\textbackslash n\textbackslash n})---the
\emph{task interface}. The \emph{negative control}
keeps the same per-prompt number of full tokens but places them at random inside the marked documents, at least 32
tokens from either end, so that the instruction, question, assistant header and prompt ends are all reduced; the
\emph{inverted control} reduces exactly the task interface, template tokens before the answer included, and keeps
the documents full. A \emph{full-budget control}
sends every token through the reduction code path with all experts, to measure the implementation's numerical
effect; an \emph{unmarked} run of the unmodified model measures the effect of the marks themselves.

\paragraph{Prespecification.}
For each of the three quality-only batches of Section~\ref{sec:q2}, the prompt manifest (hashed), the methods,
the primary and secondary comparisons and their decision rules were written into the batch's analysis script before
the formal run; we call these comparisons \emph{prespecified}. Each batch was preceded by a pilot of six prompts
per task from the same sources, used only to check that the pipeline runs and produces every output. There is no external registry or timestamp,
and later batches were designed knowing the earlier results; in particular the necessity threshold changed between
batches (Appendix~\ref{app:interf2}). The joint batch had no prespecified comparison, and comparisons we added after
seeing a batch's data are marked \emph{post hoc}.

\paragraph{Implementation.}
All rules share one vLLM hook on the fused-MoE layer. Full tokens are appended to the batch as extra rows carrying
their remaining $K-\kp$ experts, so the expert kernel is called once per layer (Supplement~S2). In the
vLLM 0.31 batches of Sections~\ref{sec:q2}--\ref{sec:q3} the service checks, on the server's own token ids, that
every prompt's reduced-token mask matches the rule, and a configuration that fails is not evaluated; earlier sessions
checked a sample of prompts (Supplement~S1 and~S9). Different results in this paper come from
different batches, engines and masks; Appendix Table~\ref{tab:evidence} lists the source of each claim. The joint
batch of Sections~\ref{sec:q2}--\ref{sec:q3} is the only one in which quality and latency share one mask, one
implementation and one set of prompts.

\paragraph{Cost.}
If $F$ of $P$ prompt tokens stay full, a rule keeps $(P-F)/P$ of the expert work that uniform reduction saves. On
the joint batch below, \ours{} keeps about 4\% of prompt tokens full and so uses 52\% of the unmodified routed-expert
slots against 50\% for uniform reduction.

\FloatBarrier
\section{Sources of Evidence}

Table~\ref{tab:evidence} lists, for each claim, the model, engine and GPU, mask and measurement it rests on.
\label{app:evidence}

\begin{table}[!htbp]
\centering\small
\caption{\textbf{Where each claim comes from.} Quality and latency come from the same mask, implementation and
prompts only in the joint batch. Other rows are separate batches, each compared with its own unmodified model.}
\label{tab:evidence}
\setlength{\tabcolsep}{4pt}
\begin{tabular}{@{}>{\raggedright\arraybackslash}p{3.6cm}>{\raggedright\arraybackslash}p{2.3cm}>{\raggedright\arraybackslash}p{2.3cm}>{\raggedright\arraybackslash}p{2.3cm}>{\raggedright\arraybackslash}p{3.4cm}@{}}
\toprule
Claim & Model & Engine, GPU & Mask & Measured \\
\midrule
Loss is form; where it is made (\S\ref{sec:analysis}) & Qwen3, DeepSeek-V2-Lite, OLMoE, Qwen1.5 & HF hooks / vLLM 0.10.2; RTX PRO 6000 and H100 (App.~\ref{sec:compute}) & unmarked templates, positional & quality \\
Protecting task tokens; equal budget (\S\ref{sec:q2}, App.~\ref{app:placement}) & Qwen3-30B-A3B & vLLM 0.31, H100 SXM & marked spans, equal-budget positions & quality, TTFT, end-to-end, throughput (joint batch) \\
Necessity of the task interface (\S\ref{sec:q2}, App.~\ref{app:placement}) & Qwen3-30B-A3B & vLLM 0.31, H100 & marked spans, negative and inverted controls & quality (negative-control batch) \\
Second placement, assistant header, DeepSeek (\S\ref{sec:q2}, App.~\ref{app:placement}) & Qwen3-30B-A3B, DeepSeek-V2-Lite & vLLM 0.31, H100 & marked spans, header and negative controls & quality (interface batch) \\
Five placements, DeepSeek screen and prefix sweep (\S\ref{sec:q2}, App.~\ref{app:placement}) & Qwen3-30B-A3B, DeepSeek-V2-Lite & vLLM 0.31, H200 & marked spans, negative controls, prefix and suffix rules & quality (placement batch) \\
Sinks, massive activations, super experts (\S\ref{sec:sinks}) & Qwen3-30B-A3B, DeepSeek-V2-Lite & HF hooks, H100 PCIe; vLLM 0.31 for the prefix sweep & uniform, positional, signal rules, restorations & answer NLL, F1, attention and activation statistics \\
Layouts, families, larger samples (Supplement~S3) & Qwen3 + three families & vLLM 0.10.2, H100 & marked spans & quality \\
Long-prompt TTFT (Supplement~S10) & Qwen3, Qwen3.6 & vLLM 0.31, H100 PCIe & marked spans & TTFT, one token \\
Qwen3.6 slowdown (Supplement~S9) & Qwen3.6-35B-A3B & vLLM 0.31, H100 & marked spans & quality; throughput, one token \\
Rerun on the same question IDs (Supplement~S9) & Qwen3-30B-A3B & vLLM 0.31, A100 & marked spans & quality, synthetic timing \\
Label-free selector (Supplement~S3.4) & four families & HF hooks & documents + whole demonstrations & quality (exploratory) \\
\bottomrule
\end{tabular}
\end{table}

\section{Joint Batch: Further Results}
\label{app:joint}

\paragraph{Cycle 1.}
Table~\ref{tab:joint-c1} gives the latency ratios of the first cycle, in which every method processes all 760
prompts once after a one-token warm-up pass over the same prompts. The unmodified model's cycle-1 throughput pass
compiled two Triton kernels during measurement (no other server did, the kernel cache being on disk by then), so
the cycle-1 throughput ratios are slightly optimistic; cycles 2 and 3 in Table~\ref{tab:joint} are clean. Every
server passed the per-prompt mask check before measurement.

\begin{table}[!htbp]
\centering\small
\caption{\textbf{Joint batch, cycle 1} (all 760 prompts): median TTFT ratio, median end-to-end ratio and
throughput ratio at concurrency four, each against the unmodified model in the same cycle.}
\label{tab:joint-c1}
\begin{tabular}{@{}lccc@{}}
\toprule
Method & TTFT & End-to-end & Throughput \\
\midrule
Unmodified, unmarked prompt & 0.987 & 0.990 & 1.011 \\
Full-budget control & 1.030 & 1.666 & 0.919 \\
Uniform & 0.889 & 0.904 & 1.221 \\
\keepends{} (32, 32) & 0.932 & 0.927 & 1.208 \\
Template role & 0.946 & 0.938 & 1.209 \\
Equal budget, ends & 0.959 & 0.955 & 1.200 \\
Equal budget, rand & 0.937 & 0.936 & 1.205 \\
Equal budget, span & 0.905 & 0.916 & 1.204 \\
\ours{} & 0.933 & 0.937 & 1.206 \\
\bottomrule
\end{tabular}
\end{table}

\begin{table}[!htbp]
\centering\small
\caption{\textbf{Cycle-1 TTFT by prompt length} (joint batch, Qwen3-30B-A3B): median per-prompt ratio to the
unmodified model. LongBench: 360 prompts, median 10.9K tokens; HotpotQA: 400 prompts, medians 1.8K and 2.3K for the
two layouts. One server start per method; no interval across starts. Throughput was not measured by length.}
\label{tab:ttft-length}
\begin{tabular}{@{}lccc@{}}
\toprule
Method & All 760 & LongBench & HotpotQA \\
\midrule
Uniform & 0.89 & 0.85 & 0.96 \\
\keepends{} (32, 32) & 0.93 & 0.86 & 1.01 \\
Template role & 0.95 & 0.86 & 1.04 \\
Equal budget, ends & 0.96 & 0.87 & 1.05 \\
Equal budget, rand & 0.94 & 0.87 & 1.02 \\
Equal budget, span & 0.90 & 0.85 & 1.00 \\
\ours{} & 0.93 & 0.86 & 1.02 \\
Full-budget control & 1.03 & 1.02 & 1.05 \\
\bottomrule
\end{tabular}
\end{table}

\paragraph{Per-task differences.}
On the 200 long-system-prompt questions, \ours{} changes F1 by $+1.63$ [$-0.79$, $+4.23$] against the unmodified
model, and is ahead of \keepends{} by $+2.17$ [$-1.22$, $+5.40$], of the template-role rule by $+1.88$
[$-1.04$, $+4.83$], and of the three equal-budget controls by $+1.51$, $+1.12$ and $+1.55$, every interval including
zero. On the 200 standard-layout questions the corresponding differences are $+1.40$ (unmodified), $+2.00$
(\keepends{}) and $+1.20$ to $+1.72$ (equal-budget controls), again all with intervals including zero.
\keepends{} loses only $-0.54$ [$-3.75$, $+2.78$] on the long-system questions here, against $-3.51$ on 2{,}000 such
questions in the earlier vLLM~0.10.2 batch. Two A100 runs with vLLM~0.31 on the same 2{,}000 questions give $-1.55$
[$-2.50$, $-0.60$] and $-1.03$ (Supplement~S9), so the loss persists on the newer engine at the larger sample
and the joint batch's estimate is best read as imprecise.

\paragraph{Marks and positions in the earlier batch.}
The larger earlier batch (vLLM 0.10.2, 2{,}000--6{,}000 questions per layout,
Supplement~S3) had no equal-budget control but points the same way. In the standard HotpotQA layout, where \ours{} keeps \V{flops.role.std.full}\
full tokens per prompt against 64 for \keepends{}, the two are indistinguishable (\V{v3pair.k4.hp.f1.full}). With a
long system prompt \ours{} wins by \V{v3pair.k4.sys.f1.full} points, but it keeps on average 609 full tokens there,
1.23 times the routed-expert slots of \keepends{} on the same prompts, because 32 head tokens cannot cover a 700-token
system prompt; the larger budget may explain the win, but that batch cannot show it. In that earlier batch \keepends{} loses 3.51 points on the 2{,}000 long-system questions, 2.47 of them
from answers that still contain the gold answer but are no longer brief (below), which is the
form loss of Section~\ref{sec:form}.
One observation is unexplained: on long system prompts the equal-budget ends control, whose head of about 300 tokens
still stops short of the instruction after a 700-token system prompt, loses little in every batch, while the fixed
32-token head of \keepends{} loses mostly form, although both keep the assistant header full. What the longer head does here is
not resolved, and it qualifies the statement that only the boundary tokens matter.

\paragraph{Form and content of the long-system loss (earlier vLLM 0.10.2 batch, not the joint batch).}
\label{app:format}
On the earlier batch (2{,}000 long-system questions), we split each method's F1 change into the questions on which
answer containment changed (content), the questions on which both runs contain the gold answer but the answer's
other words change (form), and the remainder. For \keepends{} the total of $-3.51$ is $-0.76$ content, $-2.47$ form
and $-0.28$ remainder; the share of answers that consist of the gold answer alone falls by 8.0 points. For \ours{}
the total of $-0.24$ is $-0.19$ content, $0.00$ form and $-0.05$ remainder. Containment is a lenient content test (longer answers find
the gold string more easily), so the content share may be underestimated.

\section{Negative-Control Batch}
\label{app:negctl}

\paragraph{Design.}
The batch was designed to test whether keeping the task interface full is necessary at the budget of \ours{}. Its
methods, tasks, primary comparison, decision rule and secondary comparisons were written into the analysis script
before the batch ran. Every method serves Qwen3-30B-A3B on vLLM 0.31 on one H100 (80\,GB HBM3) with greedy decoding,
reasoning off, an 8192-token step budget and full generation, once over one frozen manifest. The manifest holds 300
HotpotQA questions in each of three layouts (long system prompt, standard, question first; the same questions), 150
prompts from each of six LongBench QA sets and 150 from each of LCC and RepoBench-P. While building the manifest we
asserted, for every prompt, that the negative control keeps exactly as many full tokens as \ours{}, that none of
them lies on the task interface or within 32 tokens of either end. Two long-system prompts had fewer admissible
document positions than \ours{} full tokens (472 against 603) and were removed for every method, leaving 2{,}098.
All six reported methods completed every prompt and passed the mask check; no kernel was compiled during generation.
A second random seed of the negative control was scheduled last and was lost with the machine; it was rerun in the
interface batch (Appendix~\ref{app:interf}).

\paragraph{Analysis.}
QA datasets are equally weighted; the three HotpotQA layouts of a question are averaged into one unit, so HotpotQA
counts as one dataset and its layouts are resampled together. Intervals use 4{,}000 bootstrap draws within datasets.
The decision rule was \emph{necessary} if the primary 95\% lower bound exceeds zero and the point estimate is at
least two points (about half the gain of \ours{} over uniform reduction in the joint batch), \emph{placement does not
matter} if the 90\% interval lies within $\pm2$, and undecided otherwise.

\begin{table}[!htbp]
\centering\small
\caption{\textbf{Negative-control batch, per task}: change against the unmodified model (F1 points; edit similarity
for LCC and RepoBench-P). Descriptive only.}
\label{tab:negctl-task}
\setlength{\tabcolsep}{3pt}
\resizebox{\textwidth}{!}{\begin{tabular}{@{}lccccccccccc@{}}
\toprule
Method & HP sys & HP std & HP q-first & 2Wiki & LB HotpotQA & MuSiQue & MFQA & Narr. & Qasper & LCC & RepoB. \\
\midrule
Uniform & $-2.8$ & $-1.7$ & $-3.4$ & $-1.3$ & $+2.8$ & $-4.6$ & $-2.9$ & $-3.8$ & $-2.6$ & $-5.8$ & $-1.4$ \\
Equal budget, ends & $-0.7$ & $+1.4$ & $-1.0$ & $+3.2$ & $+1.4$ & $+1.8$ & $+0.4$ & $-1.4$ & $-0.2$ & $-2.2$ & $-0.1$ \\
Negative control & $-2.2$ & $-1.2$ & $-3.1$ & $-4.8$ & $+0.3$ & $-3.7$ & $-2.1$ & $-3.6$ & $-3.3$ & $-6.2$ & $+0.4$ \\
Inverted control & $-1.0$ & $-2.9$ & $-2.3$ & $-0.6$ & $-1.5$ & $-2.6$ & $-1.5$ & $-1.3$ & $-3.5$ & $-2.2$ & $+0.4$ \\
\ours{} & $+0.8$ & $+1.9$ & $-1.0$ & $+2.6$ & $+0.5$ & $+2.0$ & $-0.6$ & $-2.0$ & $+0.2$ & $-2.1$ & $-0.4$ \\
\bottomrule
\end{tabular}}
\end{table}

\section{Interface Batch}
\label{app:interf}

\paragraph{Design.}
The batch answers two questions left open by the negative-control batch: whether the answer interface can be
separated from the prompt ends, and whether the Qwen3 interface reduces to the assistant header. Qwen3's assistant
header is generated by its chat template and is always the last eight tokens of the prompt, so no layout moves it; we therefore
used DeepSeek-V2-Lite-Chat, whose sensitive tokens in Section~\ref{sec:analysis} are at the opening instruction, for
the first question and Qwen3-30B-A3B for the second. Both parts ran on one H100 (80\,GB) with vLLM 0.31, greedy
decoding, full generation and one pass over a frozen manifest per model, DeepSeek first; methods, primary comparisons,
decision rules and secondary comparisons were written into the analysis script before the batch ran, and a pilot of
six prompts per task preceded it on the same machine.

\emph{DeepSeek-V2-Lite} ($6\to3$): 400 HotpotQA questions in the standard layout and the same 400 with the
instruction in the middle (demonstrations, first half of the documents, instruction, second half, question). On this
layout \ours{} keeps every instruction token full and the equal-budget ends control none, which we checked with the
tokenizer before the run. Methods: unmodified, uniform, \ours{}, equal budget at the ends, negative control. Primary:
\ours{} minus the ends control on the instruction-in-the-middle layout; \emph{marks beat positions} if the 95\% lower
bound exceeds zero and the point estimate is at least two points, \emph{not distinguished} if the 90\% interval lies
within $\pm2$, undecided otherwise.

\emph{Qwen3-30B-A3B} ($8\to4$): five HotpotQA layouts (standard, long system prompt, question first, documents as tool
output, instruction in the middle) of 300 questions and six LongBench QA sets of 100; two long-system prompts were
removed for every method by the rule of Appendix~\ref{app:negctl}, leaving 2{,}098. Methods: unmodified, \ours{}, header
only, no header (Table~\ref{tab:interf}), a second seed of the negative control and equal budget at the ends; every
control matches \ours{}'s number of full tokens prompt by prompt, which was asserted while building the manifest.
Primary: \ours{} minus header only over the QA family (HotpotQA layouts pooled per question); \emph{header suffices} if the
90\% interval lies within $\pm2$, \emph{header does not suffice} if the 95\% lower bound exceeds zero and the point
estimate is at least two points, undecided otherwise.

All eleven methods completed every prompt and passed the mask check. The same-question comparison of the two
negative-control placements in Table~\ref{tab:interf} uses the 1{,}498 prompts present in both manifests (three
HotpotQA layouts and the six LongBench sets); on them the unmodified model and \ours{} return identical text in both
batches.

\paragraph{Results.}
The batch repeats the negative control with a second random
placement and probes the interface directly; its two primary comparisons, decision rules and secondary comparisons
were again fixed before it ran (Table~\ref{tab:interf}). Both primaries came out undecided.
\emph{Second placement.} The prespecified comparison gives \ours{} $+1.63$ [$-0.08$, $+3.29$] above the second
placement over this batch's tasks (Holm $p=0.18$), against $+3.22$ for the first placement on the
negative-control batch's tasks. Both batches decode greedily, and the unmodified model and \ours{} produce
identical outputs on the 1{,}498 prompts the two batches share, so the two placements can also be compared on the
same questions (post hoc): $+3.56$ [$+1.76$, $+5.40$] for the first and $+1.74$ [$+0.04$, $+3.39$] for the
second, which would not meet the two-point threshold of the primary rule. The direction replicates; the size depends
on where the random full tokens fall, and two placements do not estimate that spread (Appendix~\ref{app:interf2}
adds five).
\emph{Second family.} On DeepSeek-V2-Lite ($6\to3$, 400 HotpotQA questions in each of two layouts) the
equal-budget negative control is $4.28$ [$1.87$, $6.72$] points below \ours{} (Holm $p=0.002$). The difference comes
from the standard layout, where uniform reduction loses 6.8 points and the negative control 7.0 while \ours{} and the
equal-budget ends control lose none; it therefore does not separate the interface from the prompt ends either.
\emph{Assistant header.} On Qwen3 (2{,}098 prompts: five HotpotQA layouts of 300 questions, two long-system prompts
removed, and six LongBench sets of 100), keeping only the eight-token assistant header full and placing the rest of
\ours{}'s budget at random in the documents (\emph{header only}) is $1.01$ [$-0.61$, $+2.64$] below \ours{}; its
90\% interval [$-0.34$, $+2.34$] does not lie within the prespecified $\pm2$ margin, so the primary comparison is
undecided. Reducing the header in the \ours{} mask and moving its eight tokens of budget into the documents
(\emph{no header}) is $1.86$ [$0.38$, $3.39$] points below \ours{} (Holm $p=0.064$). Both results are compatible
with the header carrying much of the effect; neither is conclusive, and the $\pm2$ margin is about as wide as the
whole loss of uniform reduction in the negative-control batch ($2.15$), so an undecided or within-margin result here
has little power to separate rules. This batch has no uniform arm on Qwen3, so we cannot say how sensitive its
tool-output and instruction-in-the-middle layouts are; every rule is 1.6--2.3 points below the unmodified model on
the latter (Table~\ref{tab:interf-task}).
\emph{Instruction away from the ends.} To separate the interface from the prompt ends we placed DeepSeek's
instruction between two halves of the documents, where \ours{} keeps all of it full and the equal-budget ends
control none of it. The unmodified model still follows the instruction there (F1 52.8 and 4.7 words per answer,
against 47.8 and 6.8 in the standard layout), but the layout turned out insensitive to reduction: uniform reduction
changes F1 by $+1.8$ and no longer lengthens answers (3.7 words), and \ours{} minus the ends control is $-1.67$
[$-3.56$, $+0.19$], undecided (on Qwen3 the same layout gives $0.00$ [$-2.15$, $+2.16$]). With no loss to recover,
it cannot show whether marks beat positions. It also qualifies Section~\ref{sec:analysis}: the instruction is not
sensitive wherever it sits, so what matters in DeepSeek's standard layout is tied to the start of the prompt---the
instruction in that position, or other tokens the ends cover. Appendix~\ref{app:interf2} localises it to the first
8--32 tokens and finds no layout with the instruction away from the start that is sensitive to reduction.

\begin{table}[!htbp]
\centering\small
\caption{\textbf{Interface batch} (vLLM 0.31, one H100, quality only, F1 points). Unadjusted 95\% bootstrap
intervals; $p$ is Holm-adjusted within each model over the listed secondary comparisons. Both primaries are
undecided under their prespecified rules (Appendix~\ref{app:interf}). \emph{Header only}: the eight-token assistant
header full and the rest of \ours{}'s budget at random in the documents. \emph{No header}: the \ours{} mask with the
header reduced and its budget moved into the documents. The last three rows are post hoc.}
\label{tab:interf}
\begin{tabular}{@{}llc@{}}
\toprule
Comparison & Difference [95\%] & Holm $p$ \\
\midrule
\multicolumn{3}{@{}l}{\emph{DeepSeek-V2-Lite, $6\to3$, 800 prompts}} \\
\textbf{Instruction in the middle: \ours{} $-$ equal budget, ends (primary)} & $\mathbf{-1.67}$ [$-3.56$, $+0.19$] & --- \\
Standard layout: \ours{} $-$ equal budget, ends & $-0.62$ [$-2.41$, $+1.17$] & 0.53 \\
\ours{} $-$ negative control & $+4.28$ [$+1.87$, $+6.72$] & 0.002 \\
\ours{} $-$ unmodified & $+1.34$ [$-0.64$, $+3.36$] & 0.38 \\
Uniform $-$ unmodified & $-2.48$ [$-5.05$, $+0.07$] & 0.17 \\
\midrule
\multicolumn{3}{@{}l}{\emph{Qwen3-30B-A3B, $8\to4$, 2{,}098 prompts}} \\
\textbf{\ours{} $-$ header only (primary)} & $\mathbf{+1.01}$ [$-0.61$, $+2.64$] & --- \\
No header $-$ \ours{} & $-1.86$ [$-3.39$, $-0.38$] & 0.064 \\
\ours{} $-$ negative control, second seed & $+1.63$ [$-0.08$, $+3.29$] & 0.18 \\
Instruction in the middle: \ours{} $-$ equal budget, ends & $0.00$ [$-2.15$, $+2.16$] & 0.99 \\
Tool output: \ours{} $-$ header only & $-2.23$ [$-5.04$, $+0.48$] & 0.22 \\
\ours{} $-$ equal budget, ends (post hoc) & $-0.16$ [$-1.12$, $+0.83$] & --- \\
\midrule
\multicolumn{3}{@{}l}{\emph{Same 1{,}498 questions, \ours{} $-$ negative control (post hoc)}} \\
First placement (Table~\ref{tab:negctl} batch) & $+3.56$ [$+1.76$, $+5.40$] & --- \\
Second placement & $+1.74$ [$+0.04$, $+3.39$] & --- \\
\bottomrule
\end{tabular}
\end{table}

\begin{table}[!htbp]
\centering\small
\caption{\textbf{Interface batch, per task}: change against the unmodified model (F1 points). Descriptive only.
DeepSeek rows: standard / instruction-in-the-middle HotpotQA.}
\label{tab:interf-task}
\setlength{\tabcolsep}{3pt}
\resizebox{\textwidth}{!}{\begin{tabular}{@{}lccccccccccc@{}}
\toprule
Qwen3 method & HP std & HP sys & HP q-first & HP tool & HP mid & 2Wiki & LB HotpotQA & MuSiQue & MFQA & Narr. & Qasper \\
\midrule
\ours{} & $+1.9$ & $+0.8$ & $-1.0$ & $+1.8$ & $-1.8$ & $+4.2$ & $+1.5$ & $+2.3$ & $-1.2$ & $-1.4$ & $+0.4$ \\
Header only & $+0.3$ & $-0.3$ & $-0.4$ & $+4.1$ & $-1.6$ & $+3.7$ & $+3.8$ & $-2.1$ & $-2.6$ & $-2.8$ & $-1.3$ \\
No header & $-0.8$ & $+0.2$ & $-2.5$ & $-3.8$ & $-2.3$ & $+0.7$ & $+2.5$ & $-3.4$ & $-2.7$ & $-2.4$ & $+0.3$ \\
Negative control, seed 2 & $-0.6$ & $-0.5$ & $-3.3$ & $+1.8$ & $-2.0$ & $+3.5$ & $+0.8$ & $-2.3$ & $-2.0$ & $-2.2$ & $-2.3$ \\
Equal budget, ends & $+1.4$ & $-0.7$ & $-1.0$ & $+2.4$ & $-1.8$ & $+4.0$ & $+0.9$ & $+1.9$ & $+1.2$ & $-1.4$ & $+0.6$ \\
\midrule
DeepSeek method & \multicolumn{2}{c}{HP std / mid} & & & & & & & & & \\
\midrule
Uniform & \multicolumn{2}{c}{$-6.8$ / $+1.8$} & & & & & & & & & \\
Negative control & \multicolumn{2}{c}{$-7.0$ / $+1.1$} & & & & & & & & & \\
\ours{} & \multicolumn{2}{c}{$+0.9$ / $+1.7$} & & & & & & & & & \\
Equal budget, ends & \multicolumn{2}{c}{$+1.6$ / $+3.4$} & & & & & & & & & \\
\bottomrule
\end{tabular}}
\end{table}

\section{Placement Batch}
\label{app:interf2}

\paragraph{Design.}
The batch answers two questions left open by the interface batch: how large the negative control's gap is over
several random placements, and whether some layout puts DeepSeek's instruction away from the prompt ends while
remaining sensitive to reduction. It ran on one H200 (141\,GB) with vLLM 0.31, greedy decoding, full generation and
one pass per frozen manifest, after a pilot on the same machine. Methods, decision rules and the layout-selection
rule were written into the analysis scripts before the batch ran and were revised once,
also before it ran.

\emph{DeepSeek screen} ($6\to3$). Four candidate layouts of 400 HotpotQA questions each, run with the unmodified
model and uniform reduction only, put 10, 30, 100 or 300 words of unrelated text before the instruction. Only the
100- and 300-word layouts, whose instruction the equal-budget ends control does not reach (checked with the
tokenizer), were eligible for the primary comparison; the rule took the 300-word layout if it passed and otherwise
the 100-word one, where passing meant a uniform loss of at least three points with a 95\% upper bound below zero. If one was selected, the
primary comparison was \ours{} minus the ends control on 800 held-out questions of that layout.

\emph{DeepSeek main part.} Thirteen methods on 400 standard-layout questions (and the selected layout, had there
been one): unmodified, uniform, \ours{}, equal budget at the ends, two placements of the negative control, the
inverted control, the first $k\in\{1,8,32,64\}$ tokens full and the rest uniform, and the last $8$ or $32$ tokens full.

\emph{Qwen3} ($8\to4$). The interface batch's manifest of 2{,}098 prompts (identical by hash), with the unmodified
model, \ours{}, uniform, five placements of the negative control and equal budget at the ends. The primary estimate is
\ours{} minus the negative control with each question's score averaged over placements; \emph{necessary} if its
95\% lower bound exceeds zero and the point estimate is at least half of \ours{}'s gain over uniform reduction. The
placement of the interface batch was to be pooled only if the unmodified model and \ours{} reproduced at least 99\% of
that batch's outputs exactly; they reproduced 95.7\% and 96.2\% (DeepSeek: 96.5\%), which we attribute to the
change of GPU type, so it was not pooled. This concerns exact outputs, not the size of the gap: two placements on
one GPU type already differ by 1.8 points (Table~\ref{tab:interf}).

All methods completed every prompt and passed the per-prompt budget check; no prompt was removed.

\paragraph{Results.}
No candidate passed the screen (Table~\ref{tab:interf2}), so the DeepSeek primary comparison was not run. On Qwen3
the primary estimate meets the rule. The prefix and suffix rules on DeepSeek are descriptive.

\emph{Threshold.} The necessity threshold here (half of this batch's gain of \ours{} over uniform reduction, 1.30
points) differs from the negative-control batch's two points, which had been set at about half the joint batch's
gain; it was changed before this batch ran because the negative-control batch's gain of 2.61 made two points nearly
the whole gain. The new rule is defined by a quantity estimated in the same batch, so its threshold is itself
uncertain and the rule checks only the point estimate. The five-placement estimate meets either threshold; the interface-batch placement ($+1.63$) meets
neither.

\emph{Placement variance.} Seven placements of the Qwen3 negative control give point estimates from 1.6 to 3.9:
$+3.22$ (negative-control batch, H100), $+1.63$ (interface batch, H100) and $+2.33$ to $+3.88$ here (H200). On the
1{,}498 questions the two H100 batches share, where the unmodified model and \ours{} return identical text, the first
two give $+3.56$ and $+1.74$ (Table~\ref{tab:interf}), so placements on one GPU type differ by as much as question
sampling allows. The interval of the five-placement average covers question sampling only, not the choice of
placements; the H100 placements are reported separately because exact outputs do not carry over between GPU types,
not because of their size.

\begin{table}[!htbp]
\centering\small
\caption{\textbf{Placement batch} (vLLM 0.31, one H200, quality only, F1 points). Unadjusted 95\% bootstrap
intervals; $p$ is Holm-adjusted within each model over the listed secondary comparisons. Rows marked
\emph{descr.} are descriptive. \emph{First $k$ full}: the first $k$ prompt tokens at full experts, the rest uniform;
\emph{last $k$ full} likewise at the end.}
\label{tab:interf2}
\begin{tabular}{@{}llc@{}}
\toprule
Comparison & Difference [95\%] & Holm $p$ \\
\midrule
\multicolumn{3}{@{}l}{\emph{DeepSeek-V2-Lite screen, uniform $-$ unmodified, 400 questions each}} \\
10 / 30 words before the instruction & $-1.45$ [$-4.88$, $+2.01$] / $-1.87$ [$-5.10$, $+1.33$] & descr. \\
100 / 300 words before the instruction & $+0.14$ [$-3.22$, $+3.49$] / $-0.18$ [$-3.42$, $+3.06$] & --- \\
\midrule
\multicolumn{3}{@{}l}{\emph{DeepSeek-V2-Lite, standard layout, 400 questions}} \\
\ours{} $-$ negative control (two placements) & $+8.09$ [$+5.05$, $+11.26$] & $<$0.001 \\
Uniform $-$ unmodified & $-6.79$ [$-10.16$, $-3.40$] & descr. \\
First 1 / 8 full $-$ uniform & $-0.42$ [$-1.71$, $+0.90$] / $+3.80$ [$+1.88$, $+5.88$] & descr. \\
First 32 / 64 full $-$ uniform & $+8.33$ [$+5.63$, $+11.21$] / $+9.40$ [$+6.67$, $+12.19$] & descr. \\
Last 8 / 32 full $-$ uniform & $-4.06$ [$-6.72$, $-1.46$] / $-1.68$ [$-4.59$, $+1.25$] & descr. \\
Inverted control $-$ uniform & $-0.98$ [$-3.34$, $+1.30$] & descr. \\
\midrule
\multicolumn{3}{@{}l}{\emph{Qwen3-30B-A3B, $8\to4$, 2{,}098 prompts}} \\
\textbf{\ours{} $-$ negative control, five placements (primary)} & $\mathbf{+3.05}$ [$+1.47$, $+4.68$] & --- \\
\quad per placement & $+2.99$, $+3.88$, $+3.16$, $+2.33$, $+2.90$ & descr. \\
\ours{} $-$ uniform & $+2.59$ [$+0.75$, $+4.47$] & 0.022 \\
Uniform $-$ unmodified & $-1.83$ [$-3.82$, $+0.17$] & 0.22 \\
Negative control $-$ uniform & $-0.46$ [$-1.42$, $+0.45$] & 0.48 \\
\ours{} $-$ unmodified & $+0.76$ [$-0.49$, $+2.06$] & 0.48 \\
\ours{} $-$ equal budget, ends & $-0.33$ [$-1.25$, $+0.58$] & descr. \\
\bottomrule
\end{tabular}
\end{table}

\FloatBarrier
\section{Further Analysis}
\label{app:more-analysis}

Table~\ref{tab:examples} shows what a loss of form looks like in HotpotQA answers.

\begin{table}[!htbp]
\centering\small
\caption{\textbf{Reducing single spans, by family} (family batch, HotpotQA, 2{,}000 questions, half of each model's
experts; change against the unmodified model with 95\% intervals). F1, except containment for Qwen1.5, whose F1 rises
under reduction. ``Only $X$'': $X$ reduced, every other prompt token full; ``keep $X$'': $X$ full, every other prompt
token reduced. The head is the chat prefix, the instruction and the two demonstrations. Qwen3's single-token rows
come from a separate batch (Section~\ref{sec:where}).}
\label{tab:spans}
\setlength{\tabcolsep}{3pt}
\resizebox{\textwidth}{!}{\begin{tabular}{@{}lllll@{}}
\toprule
Reduced tokens & Qwen3-30B-A3B (F1) & DeepSeek-V2-Lite (F1) & Qwen1.5-MoE (containment) & OLMoE-1B-7B (F1) \\
\midrule
All prompt tokens (uniform) & \V{fam.q3.loc.all.full} & \V{fam.ds.loc.all.full} & \V{fam.q15.loc.all.hit.full} & \V{fam.olmoe.loc.all.full} \\
Only the first token & --- & \V{fam.ds.loc.first1.full} & \V{fam.q15.loc.first1.hit.full} & \V{fam.olmoe.loc.first1.full} \\
Only the first 20 & \V{fam.q3.loc.first20.full} & \V{fam.ds.loc.first20.full} & \V{fam.q15.loc.first20.hit.full} & \V{fam.olmoe.loc.first20.full} \\
Only the head & \V{fam.q3.loc.head.full} & \V{fam.ds.loc.head.full} & \V{fam.q15.loc.head.hit.full} & \V{fam.olmoe.loc.head.full} \\
Only the last two & --- & \V{fam.ds.loc.tail2.full} & \V{fam.q15.loc.tail2.hit.full} & \V{fam.olmoe.loc.tail2.full} \\
Only the last 32 & \V{fam.q3.loc.tail.full} & \V{fam.ds.loc.tail.full} & \V{fam.q15.loc.tail.hit.full} & \V{fam.olmoe.loc.tail.full} \\
Keep the last two full & \V{fam.q3.loc.keeptail2.full} & \V{fam.ds.loc.keeptail2.full} & \V{fam.q15.loc.keeptail2.hit.full} & \V{fam.olmoe.loc.keeptail2.full} \\
Keep the last 32 full & \V{fam.q3.loc.keeptail.full} & \V{fam.ds.loc.keeptail.full} & \V{fam.q15.loc.keeptail.hit.full} & \V{fam.olmoe.loc.keeptail.full} \\
Keep the first and last 32 full & \V{fam.q3.loc.keepends.full} & \V{fam.ds.loc.keepends.full} & \V{fam.q15.loc.keepends.hit.full} & \V{fam.olmoe.loc.keepends.full} \\
\bottomrule
\end{tabular}}
\end{table}

\begin{table}[!tbp]
\centering\small
\caption{\textbf{What a loss of form looks like.} HotpotQA answers of the unmodified model, with every prompt token at
half of its experts, and under \keepends{}. The reduced model still finds the answer; it no longer gives ``only the
short answer''.}
\label{tab:examples}
{\let\ours\keepends\input{generated/tab_examples.tex}}
\end{table}

\begin{figure}[!tbp]
\centering
\includegraphics[width=\textwidth]{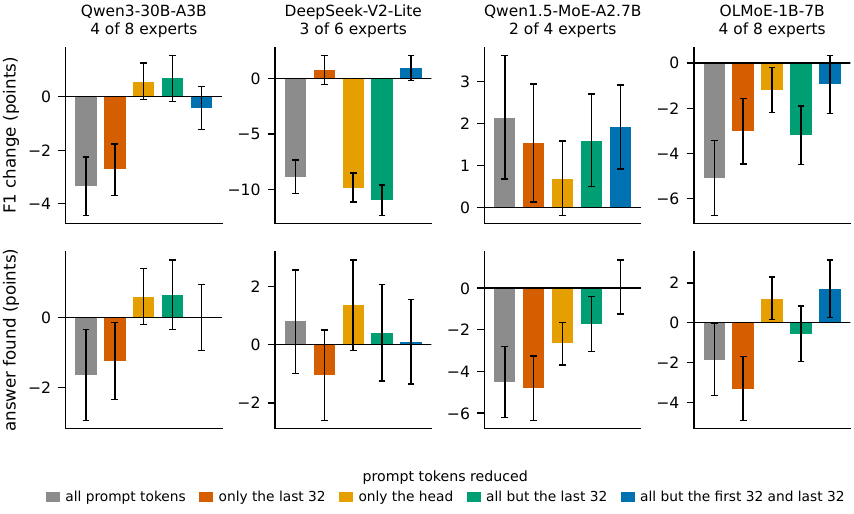}
\caption{\textbf{Where the loss is made, by model family.} HotpotQA F1 against each unmodified model when the named
prompt tokens run half of their experts (2{,}000 questions, 95\% intervals). The head is the chat prefix, the
instruction and the two demonstrations. Which end carries the loss differs by model; with 32 tokens kept at each
end (rightmost bar, the \keepends{} rule) no model shows a detectable loss on this task (Qwen1.5's F1 rises).}
\label{fig:families}
\end{figure}

\subsection{It is the answer boundary, not a template}
\label{sec:modes}

\begin{figure}[!tbp]
\centering
\includegraphics[width=\textwidth]{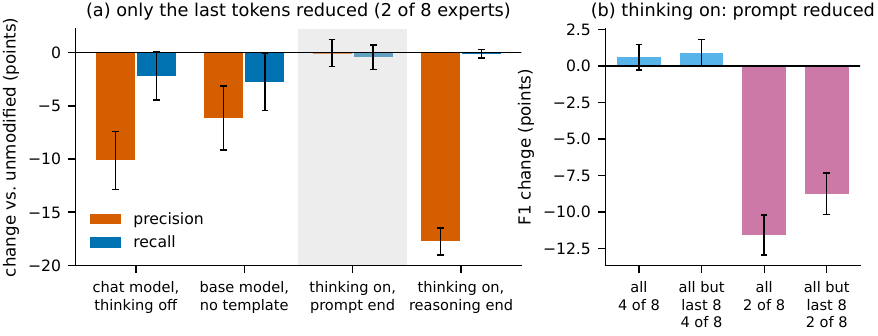}
\caption{\textbf{The effect follows the answer boundary.} (a) Reducing only the last two tokens (the last eight for
the prompt end under reasoning) to $8\to2$ in four settings: it damages precision whenever those tokens are followed directly by the answer, and not when
generated reasoning separates them from it (shaded). (b) With reasoning on, a prompt reduced to four experts is
not separated from the unmodified model by the 95\% interval in this earlier run; that alone does not establish equivalence. At two experts, prompt reduction again changes answer form.}
\label{fig:modes}
\end{figure}

\emph{Claim tested: the effect is an artefact of the template's empty reasoning block.} Three settings in
Figure~\ref{fig:modes}a say otherwise. With a \textbf{base model and no template}, where the prompt ends in
``Answer:'', reducing the last two tokens costs \V{modes.raw.f1.abs} F1 points, again mostly precision. With
\textbf{reasoning enabled}, the tokens before the answer are generated and always run all experts; reducing the end
of the prompt has no statistically detectable F1 effect in this comparison (\V{modes.think.f1.full}). But when the same reasoning trace is \textbf{replayed as
a prompt}, so that the end of the reasoning becomes reducible, reducing its last two tokens costs \V{modes.trace.f1.abs}
F1 points and \V{modes.trace.precision.abs} of precision. The sensitive tokens are whichever ones immediately precede
the answer. One consequence is practical (Figure~\ref{fig:modes}b): with reasoning on, uniform reduction of the whole prompt to
four of eight experts shows no detectable loss on this task (\V{think.direct_k4.f1.full}), which does not establish equivalence.

\subsection{Content: the question, and the paragraphs that hold the answer}
\label{sec:content}

Separate base-model interventions show sensitivity to question content. They do not fully locate the residual loss of \ours{} on marked prompts. With a base model and the
last token always full, reducing only the question (1.3\% of the prompt) lowers answer containment by
\V{hf.q_0_4.hit.abs} points, while reducing only the documents changes it by \V{hf.ctx_0_4.hit}
(Supplement~S7); a tail of 32 tokens covers the trailing question and removes most of the loss in Figure~\ref{fig:where}a.
Within the tested documents, gold paragraphs are more sensitive than the distractors: at two experts, reducing the two gold
paragraphs costs \V{docs.hotpot.gold.abs} F1 points, and reducing the eight distractors, two thirds of the prompt,
changes F1 by \V{docs.hotpot.dist.full} (Supplementary Figure~S2c). This intervention distinguishes gold from distractor paragraphs, but the serving rule has no access to gold relevance labels.

\FloatBarrier
\section{Setup, Statistics and Reproducibility}
\label{app:setup}

\paragraph{Engines.} The placement batches of Section~\ref{sec:q2} and the serving measurements of
Section~\ref{sec:q3} use vLLM 0.31 on one GPU (serving without prefix caching); every rule is a routing hook
loaded at server start and checked on every prompt against the server's token ids (Appendix~\ref{app:rules}). The
earlier unmarked analysis batches of Section~\ref{sec:analysis} use vLLM 0.10.2 with the same kind of hook
(Supplement~S2). Token-, layer- and attention-edge interventions and the direct sink tests use
HuggingFace transformers with forward hooks on the MoE blocks, eager attention and greedy decoding
(Appendix~\ref{app:sink}). The HuggingFace pipeline reproduces the engine: reducing only the last eight prompt tokens
to two experts gives \V{only.last8.f1} there and \V{form.k2.diag_inv8_k2.f1} in the engine.

\paragraph{Prompts.} HotpotQA prompts in the distractor setting hold an instruction to give only the short answer,
two demonstrations, ten paragraphs and the question, in the model's chat template with reasoning disabled; the
placement batches mark the documents with application tags (Appendix~\ref{app:rules}). GSM8K uses eight worked
examples that end with ``\#\#\#\# answer''. LongBench uses the benchmark's templates.

\paragraph{Statistics.} Every comparison is paired by question with its reference, normally the unmodified model of
the same batch. Intervals are percentile bootstraps over questions (95\% unless marked 90\%); task-averaged scores
weight tasks equally. Holm correction is applied where stated. The status of each
comparison (prespecified primary or secondary, post hoc, descriptive) follows Section~\ref{sec:background} and is
given in Table~\ref{tab:place} and the batch appendices.

\paragraph{Compute.}\label{sec:compute} All runs use single rented GPUs: an RTX PRO 6000 (96\,GB) and H100 80\,GB for
the analysis batches of Section~\ref{sec:analysis}, H100 SXM for the joint batch, H100 for the negative-control and
interface batches, H200 for the placement batch, H100 PCIe for the direct sink tests, A100-SXM4-80GB for the A100
rerun and H100 for the Qwen3.6 measurements. Table~\ref{tab:evidence} lists the engine and GPU behind each claim.

\end{document}

%% file: generated_an/numbers.tex
\expandafter\def\csname val@attn.ratio.q\endcsname{15}
\expandafter\def\csname val@budget.gsm.k2.none.abs\endcsname{14.09}
\expandafter\def\csname val@budget.gsm.k2.none.marker\endcsname{84.1}
\expandafter\def\csname val@budget.gsm.k2.t2.marker\endcsname{97.7}
\expandafter\def\csname val@budget.r_k4_t0_h8\endcsname{$-$2.78}
\expandafter\def\csname val@budget.r_k4_t0_x8\endcsname{$-$3.27}
\expandafter\def\csname val@docs.hotpot.dist.full\endcsname{$-$0.43 [$-$1.47, +0.59]}
\expandafter\def\csname val@docs.hotpot.gold.abs\endcsname{1.88}
\expandafter\def\csname val@fam.ds.loc.all.full\endcsname{$-$8.87 [$-$10.38, $-$7.37]}
\expandafter\def\csname val@fam.ds.loc.first1.full\endcsname{+0.11 [$-$0.54, +0.76]}
\expandafter\def\csname val@fam.ds.loc.first20\endcsname{$-$9.45}
\expandafter\def\csname val@fam.ds.loc.first20.full\endcsname{$-$9.45 [$-$10.67, $-$8.19]}
\expandafter\def\csname val@fam.ds.loc.head.full\endcsname{$-$9.86 [$-$11.17, $-$8.54]}
\expandafter\def\csname val@fam.ds.loc.keepends.full\endcsname{+0.95 [$-$0.17, +2.12]}
\expandafter\def\csname val@fam.ds.loc.keeptail.full\endcsname{$-$11.02 [$-$12.40, $-$9.64]}
\expandafter\def\csname val@fam.ds.loc.keeptail2.full\endcsname{$-$9.05 [$-$10.48, $-$7.62]}
\expandafter\def\csname val@fam.ds.loc.tail.full\endcsname{+0.78 [$-$0.52, +2.08]}
\expandafter\def\csname val@fam.ds.loc.tail2.abs\endcsname{3.66}
\expandafter\def\csname val@fam.ds.loc.tail2.full\endcsname{+3.66 [+2.61, +4.73]}
\expandafter\def\csname val@fam.olmoe.loc.all.full\endcsname{$-$5.11 [$-$6.76, $-$3.43]}
\expandafter\def\csname val@fam.olmoe.loc.first1.full\endcsname{$-$0.07 [$-$0.67, +0.53]}
\expandafter\def\csname val@fam.olmoe.loc.first20.full\endcsname{$-$0.69 [$-$1.52, +0.13]}
\expandafter\def\csname val@fam.olmoe.loc.head.full\endcsname{$-$1.21 [$-$2.18, $-$0.20]}
\expandafter\def\csname val@fam.olmoe.loc.keepends.full\endcsname{$-$0.94 [$-$2.25, +0.34]}
\expandafter\def\csname val@fam.olmoe.loc.keeptail.full\endcsname{$-$3.19 [$-$4.50, $-$1.90]}
\expandafter\def\csname val@fam.olmoe.loc.keeptail2.full\endcsname{$-$4.44 [$-$6.03, $-$2.80]}
\expandafter\def\csname val@fam.olmoe.loc.tail.full\endcsname{$-$3.02 [$-$4.47, $-$1.57]}
\expandafter\def\csname val@fam.olmoe.loc.tail2.full\endcsname{+0.03 [$-$0.96, +1.04]}
\expandafter\def\csname val@fam.q15.loc.all\endcsname{+2.13}
\expandafter\def\csname val@fam.q15.loc.all.hit\endcsname{$-$4.50}
\expandafter\def\csname val@fam.q15.loc.all.hit.full\endcsname{$-$4.50 [$-$6.20, $-$2.80]}
\expandafter\def\csname val@fam.q15.loc.first1.hit.full\endcsname{$-$0.95 [$-$1.60, $-$0.30]}
\expandafter\def\csname val@fam.q15.loc.first20.hit.full\endcsname{$-$1.45 [$-$2.25, $-$0.70]}
\expandafter\def\csname val@fam.q15.loc.head.hit.full\endcsname{$-$2.65 [$-$3.70, $-$1.65]}
\expandafter\def\csname val@fam.q15.loc.keepends.hit.full\endcsname{+0.05 [$-$1.25, +1.35]}
\expandafter\def\csname val@fam.q15.loc.keeptail.hit.full\endcsname{$-$1.70 [$-$3.05, $-$0.40]}
\expandafter\def\csname val@fam.q15.loc.keeptail2.hit.full\endcsname{$-$1.15 [$-$2.70, +0.40]}
\expandafter\def\csname val@fam.q15.loc.tail.hit.full\endcsname{$-$4.80 [$-$6.35, $-$3.25]}
\expandafter\def\csname val@fam.q15.loc.tail2.hit.full\endcsname{$-$4.40 [$-$5.75, $-$3.00]}
\expandafter\def\csname val@fam.q3.loc.all\endcsname{$-$3.34}
\expandafter\def\csname val@fam.q3.loc.all.full\endcsname{$-$3.34 [$-$4.45, $-$2.26]}
\expandafter\def\csname val@fam.q3.loc.first20.full\endcsname{+0.31 [$-$0.25, +0.88]}
\expandafter\def\csname val@fam.q3.loc.head.full\endcsname{+0.57 [$-$0.11, +1.27]}
\expandafter\def\csname val@fam.q3.loc.keepends.full\endcsname{$-$0.41 [$-$1.23, +0.39]}
\expandafter\def\csname val@fam.q3.loc.keeptail.full\endcsname{+0.70 [$-$0.17, +1.56]}
\expandafter\def\csname val@fam.q3.loc.keeptail2.full\endcsname{$-$1.01 [$-$2.07, +0.02]}
\expandafter\def\csname val@fam.q3.loc.tail.full\endcsname{$-$2.71 [$-$3.69, $-$1.76]}
\expandafter\def\csname val@flops.role.std.full\endcsname{121}
\expandafter\def\csname val@flops.share.2048\endcsname{57.8}
\expandafter\def\csname val@flops.share.8192\endcsname{41.7}
\expandafter\def\csname val@form.base.f1\endcsname{69.81}
\expandafter\def\csname val@form.base.words\endcsname{2.33}
\expandafter\def\csname val@form.k2.abl_t8_k2.precision\endcsname{$-$6.09}
\expandafter\def\csname val@form.k2.abl_t8_k2.recall\endcsname{$-$4.78}
\expandafter\def\csname val@form.k2.diag_inv8_k2.f1\endcsname{$-$8.05}
\expandafter\def\csname val@form.k2.diag_inv8_k2.precision.abs\endcsname{10.31}
\expandafter\def\csname val@form.k2.diag_inv8_k2.recall.abs\endcsname{0.53}
\expandafter\def\csname val@form.k4.direct_k4.precision\endcsname{$-$4.87}
\expandafter\def\csname val@form.k4.direct_k4.recall\endcsname{$-$1.09}
\expandafter\def\csname val@form.k4.direct_k4.words\endcsname{2.49}
\expandafter\def\csname val@form.n\endcsname{2,000}
\expandafter\def\csname val@hf.ctx_0_4.hit\endcsname{$-$0.1}
\expandafter\def\csname val@hf.q_0_4.hit.abs\endcsname{1.7}
\expandafter\def\csname val@knock.ko_ans_docs.stop\endcsname{0.64}
\expandafter\def\csname val@knock.ko_ans_q.stop\endcsname{0.74}
\expandafter\def\csname val@knock.ko_ans_tail2.stop\endcsname{20.70}
\expandafter\def\csname val@knock.ko_tail2_docs.first\endcsname{3.35}
\expandafter\def\csname val@knock.ko_tail2_instr.first\endcsname{6.21}
\expandafter\def\csname val@knock.ko_tail2_q.first\endcsname{0.43}
\expandafter\def\csname val@modes.raw.f1.abs\endcsname{5.69}
\expandafter\def\csname val@modes.think.f1.full\endcsname{$-$0.12 [$-$1.23, +1.01]}
\expandafter\def\csname val@modes.trace.f1.abs\endcsname{10.62}
\expandafter\def\csname val@modes.trace.precision.abs\endcsname{17.76}
\expandafter\def\csname val@only.last1.f1.abs\endcsname{0.14}
\expandafter\def\csname val@only.last2.f1\endcsname{$-$8.07}
\expandafter\def\csname val@only.last2.n\endcsname{500}
\expandafter\def\csname val@only.last2_minus_last1.f1.abs\endcsname{1.40}
\expandafter\def\csname val@only.last8.f1\endcsname{$-$8.02}
\expandafter\def\csname val@only.last8.f1.abs\endcsname{8.02}
\expandafter\def\csname val@only.last8_minus_last2.f1.abs\endcsname{0.70}
\expandafter\def\csname val@restore.band.best\endcsname{46}
\expandafter\def\csname val@tail.k4.n0\endcsname{$-$3.38}
\expandafter\def\csname val@tail.k4.n2\endcsname{$-$1.06}
\expandafter\def\csname val@tail.k4.n2.ci\endcsname{[$-$2.10, $-$0.01]}
\expandafter\def\csname val@tail.k4.n32.full\endcsname{+0.63 [$-$0.21, +1.47]}
\expandafter\def\csname val@think.direct_k4.f1.full\endcsname{+0.60 [$-$0.28, +1.47]}
\expandafter\def\csname val@v3pair.k4.hp.f1.full\endcsname{+0.14 [$-$0.21, +0.49]}
\expandafter\def\csname val@v3pair.k4.sys.f1.full\endcsname{+3.27 [+2.31, +4.26]}
\expandafter\def\csname val@v3speed.uniform.16384.gain\endcsname{+32}
\expandafter\def\csname val@v3speed.uniform.8192.gain\endcsname{+39}

%% file: generated/tab_examples.tex
\begin{tabular}{@{}>{\raggedright\arraybackslash}p{0.27\textwidth}>{\raggedright\arraybackslash}p{0.15\textwidth}>{\raggedright\arraybackslash}p{0.35\textwidth}>{\raggedright\arraybackslash}p{0.15\textwidth}@{}}
\toprule
Question & Unmodified & Uniform $8\to4$ & \ours{} \\
\midrule
\multicolumn{4}{@{}l}{\emph{Qwen3-30B-A3B}} \\
What was the nickname of the hitman hired by an Italian American Criminal Organization? & The Crowned Prince of the Philadelphia Mob & The nickname of the hitman hired by an Italian American criminal organization was ``The Crowned Prince of the Philadelphia Mob.'' & The Crowned Prince of the Philadelphia Mob \\
Which movie directed by James Gunn portrayed the fictional character, Ego the Living Planet? & Guardians of the Galaxy Vol. 2 & Answer: Guardians of the Galaxy Vol. 2 & Guardians of the Galaxy Vol. 2 \\
\midrule
\multicolumn{4}{@{}l}{\emph{DeepSeek-V2-Lite}} \\
Tysons Galleria is located in what county? & Fairfax County & Tysons Galleria is located in Fairfax County, Virginia. & Fairfax County \\
In which year was the King who made the 1925 Birthday Honours born? & 1865 & The King who made the 1925 Birthday Honours was born in 1865. & 1865 \\
\bottomrule
\end{tabular}